\documentclass[11pt, hidelinks]{article} 

\usepackage{hyperref}
\usepackage{url}

\newcommand{\arxiv}[1]{#1}
\newcommand{\nips}[1]{}

\nips{
	\usepackage[nonatbib]{nips_2017}
	
	\usepackage[utf8]{inputenc} 
	\usepackage[T1]{fontenc}    
	\usepackage{booktabs}       
	\usepackage{nicefrac}       
	\usepackage{microtype}      
}
\arxiv{
	\usepackage{fullpage}
}

\usepackage{amsfonts,epsfig,graphicx}
\usepackage{amsmath,amssymb,amsthm}

\usepackage{enumitem, comment, xifthen}

\usepackage{xcolor}

\newtheorem{theorem}{Theorem} 
\newtheorem{lemma}{Lemma}
\newtheorem{proposition}{Proposition}

\newcommand{\singularvalue}[2]{\sigma_{#1}(#2)}
\newcommand{\ones}{\ensuremath{1}}

\newcommand{\real}{\ensuremath{\mathbb{R}}}

\newcommand{\Order}{\mathcal{O}}

\newcommand{\indicator}[1]{\ensuremath{\mathbf{1}\{#1\}}}

\newcommand{\matsnorm}[2]{|\!|\!| #1 | \! | \!|_{{#2}}}
\newcommand{\opnorm}[1]{\ensuremath{\matsnorm{#1}{\mbox{\tiny{op}}}}}
\newcommand{\frobnorm}[1]{\ensuremath{\matsnorm{#1}{\mbox{\tiny{F}}}}}

\newcommand{\hausDis}{\mathcal{H}}

\newcommand{\defn}{\ensuremath{:\,=}}
\newcommand{\Lnorm}[2]{\ensuremath{\matsnorm{#1}{\mbox{\tiny{#2}}}}}
\newcommand{\inprod}[2]{\ensuremath{\langle #1 , \, #2 \rangle}}
\newcommand{\Exs}{\ensuremath{\mathbb{E}}}

\newcommand{\argmin}{\operatornamewithlimits{arg~min}}

\newcommand{\kl}[2]{\ensuremath{D_{\mathrm{KL}}(#1\|#2)}}
\newcommand{\reals}{\ensuremath{\mathbb{R}}}
\newcommand{\mprob}{\ensuremath{\mathbb{P}}}
\newcommand{\hamming}{\ensuremath{D_H}}
\newcommand{\half}{\ensuremath{{\frac{1}{2}}}}

\newcommand{\tracer}[2]{\ensuremath{\langle \!\langle {#1}, \; {#2}
\rangle \!\rangle}}

\newcommand{\upperonesmx}{J}
\newcommand{\identitymx}{I}

\newcommand{\numrows}{n}
\newcommand{\numcols}{d}

\newcommand{\plaincon}{c}
\newcommand{\ULOW}{\ensuremath{\plaincon_{\mbox{\scalebox{.5}{2}}}}}
\newcommand{\UUP}{\ensuremath{\plaincon_{\mbox{\scalebox{.5}{1}}}}}
\newcommand{\UHP}{\ensuremath{\plaincon_{\mbox{\scalebox{.5}{0}}}}}
\newcommand{\UMODEL}{\ensuremath{\plaincon_{\mbox{\scalebox{.5}{3}}}}}

\newcommand{\wtmatrix}{M}
\newcommand{\wt}{\wtmatrix}
\newcommand{\wtstar}{\wtmatrix^*}
\newcommand{\wttil}{\widetilde{\wtmatrix}}
\newcommand{\matrixset}{\mathbb{C}}
\newcommand{\diffset}{\matrixset_{\mbox{\tiny DIFF}}}
\newcommand{\diffmx}{\wtmatrix_{\mbox{\tiny DIFF}}}
\newcommand{\wthat}{\widehat{\wtmatrix}}
\newcommand{\wtLSE}{\widehat{\wtmatrix}_{\mbox{\tiny LS}}}

\newcommand{\nnset}{\matrixset_{\mbox{\tiny NR}}}
\newcommand{\permset}{\matrixset_{\mbox{\tiny PR}}}
\newcommand{\permone}{\permset(1)}

\newcommand{\pp}{\ensuremath{p_{\mathrm{obs}}}}
\newcommand{\obs}{\ensuremath{Y}}

\newcommand{\nnrank}{r}
\newcommand{\permrank}{\rho}
\newcommand{\temprank}{k}

\newcommand{\wtclosest}{\ensuremath{\wt_0}}

\newcommand{\rankonemx}[1]{Q^{#1}}
\newcommand{\lowLeft}{U}
\newcommand{\lowRight}{V}
\newcommand{\lowLeftVec}[1]{u^{#1}}
\newcommand{\lowRightVec}[1]{v^{#1}}
\newcommand{\lowLeftVecTilde}[1]{\widetilde{u}^{#1}}

\newcommand{\leftvec}{a}
\newcommand{\rightvec}{b}
\newcommand{\reg}[2][\numcols]{\frac{#2 #1 \log^{2.01} \numcols}{\pp}}
\newcommand{\fnpermrank}[1]{\overline{\permrank}(#1)}
\newcommand{\fnnnrank}[1]{\overline{\nnrank}(#1)}
\newcommand{\permstar}{\permrank^*}
\newcommand{\oracleeps}{\epsilon}

\newcommand{\frobballperm}[2]{B^{\text{P}}(#1, #2)}
\newcommand{\frobballnn}[2]{B^{\text{N}}(#1, #2)}

\newcommand{\piall}{\Pi}
\newcommand{\piset}[1][]{ \ifthenelse{\isempty{#1}}{ \widetilde{\Pi}}{\widetilde{\Pi}^{(#1)}}}
\newcommand{\sig}{\sigma}
\newcommand{\sigall}{\Sigma}
\newcommand{\sigset}[1][]{ \ifthenelse{\isempty{#1}}{ \widetilde{\Sigma}}{\widetilde{\Sigma}^{(#1)}}}
\newcommand{\pisigset}{\widetilde{\Gamma}}

\newcommand{\noise}{W}
\newcommand{\covnum}{\ensuremath{N}}
\newcommand{\metent}{\ensuremath{\log \covnum}}

\newcommand{\delcrit}{\ensuremath{\delta_c}}

\newcommand{\DelHat}{\ensuremath{\widehat{\Delta}}}

\newcommand{\packnum}{\ensuremath{\eta}}

\newcommand{\packvec}{g}
\newcommand{\packmat}{G}

\newcommand{\AuxEvent}{\ensuremath{\mathcal{A}_t}}

\newcommand{\mxtempone}{A}
\newcommand{\mxtemptwo}{B}
\newcommand{\mxdiag}{D}
\newcommand{\mxU}{U}
\newcommand{\mxV}{V}

\newcommand{\regpar}{\ensuremath{\lambda}}
\newcommand{\Treg}{\ensuremath{\widetilde{D}_{\regpar}}}
\newcommand{\wthatSVT}{\wthat_{\mbox{\scalebox{.5}{SVT}}}}

\newcommand{\convexclass}{\ensuremath{\mathbb{C}}}
\newcommand{\genclass}{\ensuremath{\mathbb{S}}}

\newcommand{\entrynorm}{q}
\newcommand{\subveclength}{\nu}

\title{{\bf{Low Permutation-rank Matrices:\\  Structural Properties
  and Noisy Completion}}}

\author{
	{\large{
			\begin{tabular}{ccc}
				Nihar B. Shah & Sivaraman Balakrishnan & Martin
				J. Wainwright \\
				\small ML and CS Depts. & \small  Dept. of
				Statistics & \small Dept. of EECS and
				Statistics\\ \small CMU & \small CMU & \small UC
				Berkeley \\ \small nihars@cs.cmu.edu & \small
				siva@stat.cmu.edu & \small wainwrig@berkeley.edu
			\end{tabular}
	}}
}

\date{}

\begin{document}

\maketitle


\begin{abstract}
We consider the problem of noisy matrix completion, in which the goal
is to reconstruct a structured matrix whose entries are partially
observed in noise.  Standard approaches to this underdetermined
inverse problem are based on assuming that the underlying matrix has
low rank, or is well-approximated by a low rank matrix.  In this
paper, we propose a richer model based on what we term the
``permutation-rank'' of a matrix.  We first describe how the classical
non-negative rank model enforces restrictions that may be undesirable
in practice, and how and these restrictions can be avoided by using
the richer permutation-rank model.  Second, we establish the minimax
rates of estimation under the new permutation-based model, and prove
that surprisingly, the minimax rates are equivalent up to logarithmic
factors to those for estimation under the typical low rank
model. Third, we analyze a computationally efficient
singular-value-thresholding algorithm, known to be optimal for the
low-rank setting, and show that it also simultaneously yields a
consistent estimator for the low-permutation rank setting. Finally, we
present various structural results characterizing the uniqueness of
the permutation-rank decomposition, and characterizing convex
approximations of the permutation-rank polytope.
\end{abstract}



\section{Introduction}

In the problem of matrix completion, the goal is to reconstruct a
matrix based on observations of a subset of its entries~\cite{Lau01}.
Matrix completion has a variety of applications, including recommender
systems~\cite{koren2009matrix}, image
understanding~\cite{lee1999learning}, credit risk
monitoring~\cite{vandendorpe2008parameterization}, fluorescence
spectroscopy~\cite{gobinet2004application}, and modeling
signal-adaptive audio effects~\cite{sarver2011application}. We refer
the reader to the surveys~\cite{gillis2014and,davenport2016overview}
for an overview of the vast literature on this topic.  Throughout this
paper, in order to provide a running example for our modeling, it will
be convenient to refer back to a particular variant of a recommender
system application.  More concretely, suppose that there are $\numrows
\geq 2$ users and $\numcols \geq 2$ items, as well as an unknown
matrix $\wtstar \in [0,1]^{\numrows \times \numcols}$ that captures
the users' preferences for the items. Specifically, the $(i,j)^{th}$
entry of $\wtstar$ represents the probability that user $i$ likes item
$j$. The problem is to estimate this preference matrix $\wtstar \in
[0,1]^{\numrows \times \numcols}$ from observing users' likes or
dislikes for some subset of the items.

Following a long line of past work in this area
(e.g.,~\cite{chen13,gross11,srebro2005generalization,
  candes09exact,candes10power, keshavan2010matrix, recht2011simpler,
  chatterjee2014matrix}), we consider the following form of random
design observation model.  For a given parameter $\pp
\in (0,1]$ and for any user-item pair $(i,j)$, we observe $i$'s rating
  for item $j$ with probability $\pp$. We assume that when an entry
  $(i,j)$ is observed, we observe a binary value---for instance,
  $\{$like, dislike$\}$ or $\{0,1\}$---which arises as a Bernoulli
  realization of the true preference $\wtstar_{ij}$.\footnote{Our
    results readily extend to any rating scheme with bounded values,
    such as five-star ratings. We focus on the binary case for
    purposes of brevity.}  More formally, we observe a matrix $\obs
  \in \{0,\half,1\}^{\numrows \times \numcols}$, where
\begin{align}
\label{EqnDefnObsLowrank}
\obs_{ij} =
\begin{cases}
1 \quad \mbox{with probability } \pp \wtstar_{ij} & \quad (\mbox{user
  $i$ likes item $j$}) \\
0 \quad \mbox{with probability } \pp (1-\wtstar_{ij}) & \quad
(\mbox{user $i$ dislikes item $j$})\\
\half \quad \mbox{with probability } 1 - \pp & \quad (\mbox{no data
  available}),
\end{cases}
\end{align}
for every $(i,j) \in [\numrows] \times [\numcols]$. The goal is to
estimate the underlying matrix $\wtstar$ based on the observed matrix
$\obs$.

It is clear that, if no structural conditions are imposed on the
underlying matrix $\wtstar$, then this problem is ill-posed.  A
classical approach is to impose on a bound on either the rank or the
non-negative rank of the matrix.  We begin by describing the approach
based on the non-negative rank, before turning to the alternative
approach based on permutation rank that is the focus of this paper.

\paragraph*{Non-negative rank:} In the problem of non-negative
low-rank matrix completion, the matrix $\wtstar$ is assumed to have a
factorization of the form
\begin{align*}
\wtstar = \lowLeft \lowRight^T,
\end{align*}
for some matrices $\lowLeft \in \reals_+^{\numrows \times \nnrank}$
and $\lowRight \in \reals_+^{\numcols \times \nnrank}$. Here the
integer $\nnrank \in \{1,\ldots, \min\{\numcols, \numrows\}\}$ is
known as the \emph{non-negative rank} of the matrix. (As a corner
case, we also have that the zero matrix is the one and only matrix
with a non-negative rank of $\nnrank = 0$.)  It is often assumed that
the non-negative rank $\nnrank$ is a known quantity, but in this
paper, we make no such assumptions. For any value of $\nnrank \in
\{1,\ldots,\min\{\numcols, \numrows\}\}$, we let $\nnset(\nnrank)$
denote the set of all matrices with a non-negative factorization of
rank at most $\nnrank$---that is
\begin{align*}
\nnset(\nnrank) \defn \left \{ \wtmatrix \in [0,1]^{\numrows \times \numcols} \mid \wtmatrix = \lowLeft \lowRight^T,~ \lowLeft \in \reals_+^{\numrows \times \nnrank},~ \lowRight \in \reals_+^{\numcols \times \nnrank} \right \}.
\end{align*}
For any matrix $\wtmatrix$, the smallest value of $\nnrank$ such that $\wtmatrix \in \nnset(\nnrank)$ is termed its \emph{non-negative rank},
and is denoted by $\fnnnrank{\wt}$.

In order to gain some intuition for the meaning of the non-negative rank,
note that any matrix $\wtmatrix \in \nnset(\nnrank)$ can be written as
a sum of the form
\begin{align*}
\wtmatrix = \sum_{\ell=1}^{\nnrank} \lowLeftVec{\ell} (\lowRightVec{\ell})^T.
\end{align*}
Here $\lowLeftVec{\ell} \in \reals_+^\numrows$ and $\lowRightVec{\ell}
\in \reals_+^{\numcols}$ are vectors such that $\lowLeftVec{\ell}
(\lowRightVec{\ell})^T \in [0,1]^{\numrows \times \numcols}$ for every
$\ell \in [\nnrank]$. Such a decomposition can be interpreted as the
existence of $\nnrank$ features, indexed by $\ell \in [\nnrank]$.  The
$\numcols$ entries of vector $\lowRightVec{\ell}$ represent the
contribution of feature $\ell$ to the $\numcols$ respective items, and
the $\numrows$ entries of vector $\lowLeftVec{\ell}$ represent the
amounts by which the $\numrows$ respective users are influenced by
feature $\ell$. The popular overview article by Koren et
al.~\cite{koren2009matrix} provides an explanation for this
assumption:
\begin{quote}
  \emph{ {``Latent factor models are an alternative approach that
      tries to explain the ratings by characterizing both items and
      users on, say, 20 to 100 factors inferred from the ratings
      patterns. ...  For movies, the discovered factors might measure
      obvious dimensions such as comedy versus drama, amount of
      action, or orientation to children; less well-defined dimensions
      such as depth of character development or quirkiness; or
      completely uninterpretable dimensions. For users, each factor
      measures how much the user likes movies that score high on the
      corresponding movie factor.''}  }
\end{quote}


\paragraph{Concerns with non-negative rank:}  
Let us now delve deeper into this model\arxiv{\footnote{A slightly
    different, alternative interpretation is discussed in
    Appendix~\ref{SecAlternativeInterpretationOfNNRank}.}}  continuing
in the context of movie recommendations for the sake of
concreteness. Suppose there are $\nnrank$ features that govern the
movie watching experience; examples of such features include the
amount of comedy content or the depth of character development. For
any user $i \in [\numrows]$ and any feature $\ell \in [\nnrank]$, we
let $\lowLeftVec{\ell}_i \in \reals_+$ denote the ``affinity'' of user
$i$ towards feature $\ell$, and for any movie $j \in [\numcols]$, we
let $\lowRightVec{\ell}_j \in \reals_+$ denote the amount of content
associated to feature $\ell$ in movie $j$. The conventional low
non-negative rank model then assumes that the affinity of user $i$
towards movie $j$ conditioned on feature $\ell$ is given by
$\lowLeftVec{\ell}_i \lowRightVec{\ell}_j$.  Consequently, for given
feature $\ell$, the entire behavior of each user and movie is governed
by a pair of parameters, namely $\lowLeftVec{\ell}_i$ and
$\lowRightVec{\ell}_j$ for user $i$ and item $j$ respectively. Such an
assumption has some unnatural implications. For instance, consider any
two movies, say $A$ and $B$, and any two users, say $X$ and $Y$. Then
conditioned on any feature $\ell$, we have the implication
\begin{align*}
\frac{\text{Preference of user $X$ for movie $A$}}{\text{Preference of user $X$ for movie $B$}} = \frac{\text{Preference of user $Y$ for movie $A$}}{\text{Preference of user $Y$ for movie $B$}}.
\end{align*}
In words, the low non-negative rank model inherently imposes a
condition that is potentially unrealistic---namely, that for any given
feature, the ratio of preferences for any pair of movies is identical
for all users. Likewise, for any given feature, the ratio of
preferences of any pair of users is identical for all movies.  With
the goal of circumventing this possibly troublesome condition, let us
now describe a generalization that we call permutation rank; it is the
main focus of our paper.


\paragraph*{Permutation rank:}

As with the ordinary rank, the permutation rank of the all-zeros
matrix is zero.  Otherwise, for any non-zero matrix, the permutation
rank $\permrank$ takes values in the set $\{1, \ldots, \min\{\numrows,
\numcols\}\}$.  We begin by describing the set $\permset(1)$ of
matrices with permutation rank one, before describing how to extend to
arbitrary $\permrank > 1$. The set of matrices with permutation rank
$\permrank = 1$ is given by
\begin{align*}
\permset(1) \defn & \{ \wtmatrix \in [0,1]^{\numrows \times \numcols} \mid \mbox{$\exists$ permutations $\pi_1:[\numrows] \rightarrow [\numrows]$ and $\pi_2:[\numcols] \rightarrow [\numcols]$ such that}\\
& \mbox{ $\wtmatrix_{ij} \geq \wtmatrix_{i' j'}$ for every quadruple $(i,j , i', j')$ such that $\pi_1(i) \geq \pi_1(i')$ and $\pi_2(j) \geq \pi_2(j')$ } \}.
\end{align*}
In words, a non-zero matrix is said to have a permutation rank of $1$
if there exists a permutation of its rows and columns such that the
entries of the resulting matrix are non-decreasing down any column and
to the right along any row. Observe that any matrix with the
conventional (non-negative) rank equal to $1$ also belongs to the set
$\permset(1)$. However, a matrix in $\permset(1)$ can have any
non-negative rank, meaning the set of matrices with a permutation-rank
of $1$ also includes some matrices with a full non-negative rank.

We now extend the definition of the permutation rank to any integer
$\permrank \in \{1,\ldots,\min\{\numrows, \numcols\}\}$.  In
particular, the set of matrices with permutation rank $\permrank$ is
given by
\begin{align*}
  \permset(\permrank) \defn \Big\{ \wtmatrix \in [0,1]^{\numrows\times
    \numcols} \ \Bigr\rvert \ & \mbox{$\wtmatrix =
    \sum_{\ell=1}^{\permrank} \rankonemx{\ell}$ for some matrices
    $\rankonemx{1},\ldots,\rankonemx{\permrank} \in \permset(1)$}
  \Big\},
\end{align*}
of matrices having a permutation-rank at most $\permrank$. Note that
this definition reduces to our previous one in the special case
$\permrank = 1$.  Otherwise, for $\permrank > 1$, the permutations
defining membership of each constituent matrix $\rankonemx{\ell}$ in
$\permset(\permrank)$ are allowed to be different. For any matrix
$\wtmatrix$, the smallest value of $\permrank$ such that $\wtmatrix
\in \permset(\permrank)$ is termed its \emph{permutation rank}, and is
denoted by $\fnpermrank{\wtmatrix}$.

Revisiting the example of movie recommendations, the interpretation of
this more general permutation-rank model is that conditioned on any
feature $\ell \in [\nnrank]$, the preference ordering across movies
continues to be consistent for different users, but the values of
these preferences \emph{need not} be identical scalings of each
other. Observe that the conventional non-negative matrix-completion
setting $\nnset(\nnrank)$ is a special case of the permutation-rank
matrix-completion setting where each matrix $\rankonemx{\ell}$ is
restricted to be of rank one. Whenever \mbox{$\nnrank <
  \min\{\numcols, \numrows\}$,} we have the strict inclusion
\mbox{$\nnset(\nnrank) \subset \permset(\nnrank)$.}


\paragraph*{Outline and main contributions:}

Having discussed the limitations of the non-negative rank model and
introduced the permutation rank, we now outline the remainder of the
paper. In Section~\ref{SecEstimation}, we present our main results on
the problem of estimating the matrix $\wtstar$ (in the Frobenius norm)
from partial and noisy observations. Specifically, we present a
certain regularized least squares estimator, and prove that it
achieves (nearly) minimax-optimal rates for estimation over the
permutation-rank model. We also show that surprisingly, even if one
considers the more restrictive non-negative rank model, and even if
the rank is known, no estimator can achieve lower estimation error up
to logarithmic factors. We also analyze the computationally efficient
Singular Value Thresholding (SVT) algorithm, and show that it yields
consistent estimates over the permutation-rank model, in addition to
yielding the optimal estimate under the non-negative rank model. In
Section~\ref{SecPropPermModel}, we establish some interesting
properties of the permutation-rank model, and also derive certain
relationships of this model with the non-negative rank
model. \arxiv{In Section~\ref{SecProofs} we present the proofs of our
  results.} We conclude the paper with a discussion in
Section~\ref{SecConclusion}.  \nips{ The proofs of our theoretical
  results are presented in the appendix.  }

\arxiv{ The paper also contains two
  appendices. Appendix~\ref{SecAlgoFail} is devoted to negative
  results, where we show that certain intuitive algorithms provably
  fail. Appendix~\ref{SecAlternativeInterpretationOfNNRank} describes
  an alternative interpretation of the non-negative rank model.  }


\section{Main results on estimating $\wtstar$}
\label{SecEstimation}

We begin by considering the problem of estimating a low permutation
rank matrix $\wtstar$ based on noisy and partial observations.  We
first analyze a computationally expensive estimator, based on
regularizing the least-squares cost with a multiple of the permutation
rank, and show that it achieves minimax-optimal rates up to
logarithmic factors.  We then turn to a polynomial-time algorithm
based on nuclear norm regularization, which is equivalent singular
value thresholding in the current set-up.


\subsection{Optimal oracle inequalities for estimation}
\label{SecStat}

Suppose that we collect an observation matrix $\obs$ of the
form~\eqref{EqnDefnObsLowrank}, where the unknown matrix $\wtstar$
belongs $\permset$.  In this section, we analyze a regularized form of
least-squares estimation, as applied to the recentered matrix
\begin{subequations}
\begin{align}
\label{EqnDefnObsPartialLowRank}
\obs' \defn \frac{1}{\pp} \obs - \frac{1 - \pp}{2\pp} \ones \ones^T.
\end{align}
We perform this recentering in order to obtain an unbiased estimate
($\obs'$) of the true matrix $\wtstar$ in the presence of missing
observations, which is used in the least-squares estimator described
below. As a sanity check, observe that when $\pp = 1$, we have the
direct relation $\obs' = \obs$.

Letting $\fnpermrank{\wtmatrix}$ denote the permutation-rank of any
matrix $\wtmatrix$, we then consider the estimator
\begin{align}
\label{EqnDefnLSEPartialLowRank}
\wtLSE ~\in~ \argmin \limits_{\wtmatrix \in [0,1]^{\numrows \times
    \numcols}}~ \Big( \frobnorm{ \obs' - \wtmatrix}^2 +
\reg[\max\{\numrows,\numcols\}]{\fnpermrank{\wtmatrix}} \Big).
\end{align}
\end{subequations}
Observe that importantly, the estimator $\wtLSE$ does \emph{not} need
to know the value of the true permutation-rank of the underlying
matrix. Moreover, while the estimator (as stated) is based on a known
value of $\pp$, this assumption is not critical, since $\pp$ can be
estimated accurately from the observed matrix $\obs$.

We now turn to some theoretical guarantees on the performance of this
estimator.  Rather than assuming that, for a given rank $\permrank$,
the target matrix $\wtstar$ has permutation rank exactly equal to
$\permrank$, we instead provide bounds that depend on distances to the
set of all matrices with a given permutation rank.  More precisely,
for any given tolerance $\oracleeps \geq 0$, define the set
\begin{align*}
\frobballperm{\permrank}{\oracleeps} \defn \big\{ \wtmatrix \in
             [0,1]^{\numrows \times \numcols} \mid \mbox{$\exists
               \wtmatrix' \in [0,1]^{\numrows \times \numcols}$ s.t.
               $\fnpermrank{\wt'} \leq \permrank$ and $
               \frobnorm{\wtmatrix - \wtmatrix'} \leq \oracleeps$}
             \big \},
\end{align*}
corresponding to the set of all matrices that are at most $\oracleeps$
distant from the set of matrices with permutation rank $\permrank$.
Similarly, we define the set
\begin{align*}
\frobballnn{\nnrank}{\oracleeps} \defn \big\{ \wtmatrix \in
           [0,1]^{\numrows \times \numcols} \mid \mbox{$\exists
             \wtmatrix' \in [0,1]^{\numrows \times \numcols}$ s.t.
             $\fnnnrank{\wt'} \leq \nnrank$ and $ \frobnorm{\wtmatrix
               - \wtmatrix'} \leq \oracleeps$} \big \},
\end{align*}
corresponding to matrices that are at most $\oracleeps$ away from some
matrix with non-negative-rank $\nnrank$.

In stating the following theorem, as well as throughout the remainder
of the paper, we use $\plaincon,\plaincon',\plaincon_1$ etc. to denote
positive universal constants. The values of these constants may differ
from line to line.
\begin{theorem}
  \label{ThmStatOracle}
  (a) For any matrix $\wtstar \in [0,1]^{\numrows \times \numcols}$
  and any integer $\permrank \in [\min\{\numrows,\numcols\}]$, the
  regularized least squares estimator $\wtLSE$ satisfies the upper
  bound
\begin{subequations}
  \begin{align}
    \label{EqnOracleUpper}
  \frac{1}{\numrows \numcols} \frobnorm{\wtLSE - \wtstar}^2 & \leq
  \UUP \min \Big\{ 1, \quad \min_{\wt \in \permset(\permrank)}
  \frac{\frobnorm{\wt - \wtstar}^2}{\numrows \numcols} +
  \frac{\permrank \ \log^{2.01} (\numrows \numcols )}{ \min
    \{\numrows, \numcols \} \pp } \Big\},
\end{align}
with probability at least $1 - e^{- \UHP \max\{\numrows,\numcols\}
  \log (\numrows \numcols)}$.\\
\noindent (b) Conversely, for any integer $\permrank \in [\min
  \{\numrows,\numcols\}]$, any scalar $\oracleeps \geq 0$, and any
estimator $\wthat$, there exists a matrix $\wtstar \in
\frobballnn{\permrank}{\oracleeps}$ such that
\begin{align}
  \label{EqnOracleLower}
  \Exs \big[ \frac{1}{\numrows \numcols} \frobnorm{\wthat - \wtstar}^2
    \big] \geq \ULOW \min \Big\{ 1, \quad \frac{ \oracleeps^2
  }{\numrows \numcols} + \frac{\permrank}{ \min \{\numrows, \numcols
    \} \pp } \Big\}.
\end{align}
\end{subequations}
\end{theorem}
\noindent See Section~\ref{SecProofThmStatOracle} for the proof of
this claim.

\paragraph{Interpretation as oracle inequality:}
The upper bound~\eqref{EqnOracleUpper} is an instance of an
\emph{oracle inequality}: it provides a family of upper bounds, one
for each choice of the integer $\permrank \in
[\min\{\numrows,\numcols\}]$, on the estimation error associated with
an arbitrary matrix $\wtstar \in [0,1]^{\numrows \times \numcols}$.
For each choice of $\permrank$, the upper bound~\eqref{EqnOracleUpper} consists of two terms.
The first term, involving the minimum over $\wt \in
\permset(\permrank)$, is a form of approximation error: it measures
how well the unknown matrix $\wtstar$ can be approximated with a
matrix $\wt$ of permutation rank at most $\permrank$.  The second term
is a form of estimation error, measuring the difficulty of estimating
a matrix that has permutation rank at most $\permrank$.  Since one
such an upper bound holds for each choice of $\permrank$, the
bound~\eqref{EqnOracleUpper} shows that the estimator mimicks the
behavior of an ``oracle'', which is allowed to choose $\permrank$ so
as to optimize the trade-off between the approximation and estimation
error.

\paragraph{Sandwiching of the risk:} 
The upper bound~\eqref{EqnOracleUpper} of Theorem~\ref{ThmStatOracle} can equivalently be stated in the following manner.  For any integer $\permrank \in [\max\{\numrows,\numcols\}]$ and
any scalar $\epsilon \geq 0$ such that $\wtstar \in \frobballperm{\permrank}{\oracleeps}$, the regularized least squares estimator $\wtLSE$ satisfies the upper
bound
\begin{subequations}
\begin{align}
\label{EqnOracleUpperRewrite}
\frac{1}{\numrows \numcols} \frobnorm{\wtLSE - \wtstar}^2
& \leq \UUP \min \Big\{ \frac{\oracleeps^2}{\numrows \numcols} +  \frac{\permrank \ \log^{2.01} (\numrows \numcols )}{ \min \{\numrows, \numcols \} \pp }  , 1 \Big\},
\end{align}
with probability at least $1 - e^{- \UHP  \max\{\numrows,\numcols\} \log (\max\{\numrows \numcols\})}$. 
On the other hand, since $\frobballnn{\permrank}{\oracleeps} \subseteq \frobballperm{\permrank}{\oracleeps}$ for every value of $\permrank$ and $\oracleeps$, the lower bound~\eqref{EqnOracleLower} of Theorem~\ref{ThmStatOracle} implies the following result. For any integer $\permrank \in [\min
\{\numrows,\numcols\}]$, any scalar $\oracleeps \geq 0$, and any
estimator $\wthat$, there exists a matrix $\wtstar \in
\frobballnn{\permrank}{\oracleeps}$ such that
\begin{align}
\label{EqnOracleLowerRewrite}
\Exs \big[ \frac{1}{\numrows \numcols} \frobnorm{\wthat - \wtstar}^2
\big] \geq \ULOW \min \Big\{ 1, \quad \frac{ \oracleeps^2
}{\numrows \numcols} + \frac{\permrank}{ \min \{\numrows, \numcols
	\} \pp } \Big\}.
\end{align}
\end{subequations}
Comparing the bounds~\eqref{EqnOracleUpperRewrite} and~\eqref{EqnOracleLowerRewrite}, we see that
our results are sharp up to logarithmic factors.


\paragraph{Specialization to minimax risk:}
When suitably specialized to matrices that have some fixed permutation
(or non-negative) rank, Theorem~\ref{ThmStatOracle} leads to sharp upper and
lower bounds on the minimax risks for the problems of matrix
completion over the sets $\nnset$ and $\permset$. 
In order for a clear comparison between the two bounds, let us index both the non-negative rank and the permutation-rank using a generic notation $\temprank$---the meaning of the notation will be clear from the context. 

Part
(a) of Theorem~\ref{ThmStatOracle} implies that for any value 
$\temprank \in [\min\{\numrows, \numcols\}]$ and any matrix $\wtstar
\in \permset(\temprank)$, the regularized least squares estimator
$\wtLSE$ satisfies the bound
\begin{align*}
  \frac{1}{\numcols \numrows} \frobnorm{\wtLSE - \wtstar}^2 \leq \UUP
  \min \Big\{ \frac{\temprank \ \log^{2.01} (\numrows \numcols)}{\min
    \{\numrows, \numcols \} \pp} , 1 \Big\},
\end{align*}
with probability at least $1 - e^{- \UHP \max\{\numrows,\numcols\} \log
  (\numrows \numcols)}$. Within the set of matrices $[0,1]^{\numrows \times \numcols}$ under consideration, we have the deterministic upper
bound $\frac{1}{\numrows \numcols} \frobnorm{\wtLSE - \wtstar}^2 \leq
1$. Consequently, our high probability upper bound also implies a uniform bound on the
mean-squared error over the set $\permset(\temprank)$---that is
\begin{subequations}
\label{EqnMinimax}
\begin{align}
\sup_{\wtstar \in \permset(\temprank)} \frac{1}{\numcols \numrows}
\Exs [ \frobnorm{\wtLSE - \wtstar}^2 ] \leq \UUP' \min \Big\{
\frac{\temprank \ \log^{2.01} (\numrows \numcols)}{\min \{\numrows,
  \numcols \} \pp} , 1 \Big\}.
\end{align}
Since $\permset(\temprank)$ is a superset of $\nnset(\temprank)$, the
same upper bound holds for the minimax risk over $\nnset(\temprank)$:
\begin{align}
\sup_{\wtstar \in \nnset(\temprank)} \frac{1}{\numcols \numrows}
\Exs [ \frobnorm{\wtLSE - \wtstar}^2 ] \leq \UUP' \min \Big\{
\frac{\permrank \ \log^{2.01} (\numrows \numcols)}{\min \{\numrows,
	\numcols \} \pp} , 1 \Big\}.
\end{align}

Conversely, part (b) of Theorem~\ref{ThmStatOracle} implies that for
any $\temprank \in [\max\{\numrows,\numcols\}]$, the error incurred by any estimator $\wthat$ over the set $\nnset(\temprank)$ is error lower bounded as
\begin{align}
\sup_{\wtstar \in \nnset(\temprank)} \frac{1}{\numcols\numrows} \Exs[
  \frobnorm{\wthat - \wtstar}^2 ] \geq \ULOW \min \Big\{
\frac{\temprank}{\min \{\numrows, \numcols \} \pp} , 1 \Big\}.
\end{align}
Since $\nnset(\temprank) \subseteq \permset(\temprank)$, the error incurred by any estimator $\wthat$ over the set $\permset(\temprank)$ is also lower bounded as
\begin{align}
\sup_{\wtstar \in \permset(\temprank)} \frac{1}{\numcols\numrows} \Exs[
\frobnorm{\wthat - \wtstar}^2 ] \geq \ULOW \min \Big\{
\frac{\temprank}{\min \{\numrows, \numcols \} \pp} , 1 \Big\}.
\end{align}
\end{subequations}

We have thus characterized the minimax risk over both the families
$\permset$ or $\nnset$, with bounds~\eqref{EqnMinimax} that are matching up to
logarithmic factors. An important consequence of our oracle and
minimax results is the multi-fold benefit of moving from the
restrictive non-negative-rank assumptions to the more general
permutation-rank assumptions. Fitting a permutation-rank $\temprank$
model when the true matrix actually has a non-negative rank of
$\temprank$ leads to relatively little additional (overfitting)
error. On the other hand, we show later in the paper that fitting a
non-negative rank $\temprank$ model when the true matrix actually has
a permutation-rank of $\temprank$ can lead to very high error, due to
model mismatch.

\paragraph{Link to past work:}
A special case of our present problem is equivalent to the setting
considered in \arxiv{our earlier work}\nips{the
  paper}~\cite{shah2015stochastically}, corresponding to the case when
the value of $\permrank$ is known and equal to $1$, the matrix
$\wtstar$ is square with $\numrows = \numcols$, and all entries of
$\wtstar$ satisfy the shifted-skew-symmetry condition $\wtstar_{ij} +
\wtstar_{ji} = 1$. The proof of the upper bound of
Theorem~\ref{ThmStatOracle}(a) relies on a framework laid out in this
earlier work~\cite{shah2015stochastically}, but augments the results
obtained in this past work, both in generalizing to broader families
of matrices, but also providing oracle inequalities that allow for
matrices that need not be exactly low rank (in either the non-negative
or permutation senses). Theorem~\ref{ThmStatOracle} provides
guarantees on the estimation of ``mixtures'' of different
permutations, thereby resolving an important open problem in the
paper~\cite{shah2015stochastically}.


\subsection{Computationally efficient estimator}

At this point, we do not know how to compute the regularized least
squares estimator~\eqref{EqnDefnLSEPartialLowRank} in an efficient
manner, and we suspect that it may be computationally intractable to do
so. Consequently, in this section, we turn to analyzing a different
method based on singular value thresholding (SVT). Singular value
thresholding has been used either directly or as a subroutine in
several past papers on the conventional low-rank matrix completion
problem (see, for example, the
papers~\cite{cai2010singular,donoho2014minimax,chatterjee2014matrix}).
This approach is appealing due to its computational simplicity,
involving only computation of the singular value decomposition,
followed by a single pointwise non-linearity; see Cai et
al.~\cite{cai2010fast} for a fast algorithm.  In the context of the
permutation rank completion problem, we show here that the SVT
estimator is consistent for estimation under the permutation-rank
model, albeit with a rate that is suboptimal by a factor of
$\sqrt{\min\{\numrows,\numcols\} \pp}$.  Note that these guarantees
hold without the estimator needing to know that the matrix is drawn
from a permutation-rank model, nor the value $\pp$ of the permutation
rank. \\

The SVT estimator is straightforward to describe.
From the observation matrix $\obs \in \{0,\frac{1}{2},1\}^{\numrows
  \times \numcols}$, we first obtain the transformed observation
matrix $\obs'$ as in equation~\eqref{EqnDefnObsPartialLowRank}.
Applying the singular value decomposition yields the representation
\mbox{$\obs' = U D V^T$,} where the $(\numrows \times \numcols)$
matrix $D$ is diagonal, whereas the $(\numrows \times \numcols)$
matrices $U$ and $V$ are orthonormal. For a threshold $\regpar > 0$ to
be specified, define another diagonal matrix $\Treg$ with entries
\begin{align}
\label{EqnDefnSoftSVT}
[\Treg]_{jj} & = 
\begin{cases}
0 & \quad \mbox{if $D_{jj} < \regpar$} \\
D_{jj} - \regpar  & \quad \mbox{if $D_{jj} \geq \regpar$}
\end{cases}
\qquad \mbox{for each $j \in [\max\{\numrows, \numcols\}]$.}
\end{align}
Finally, the SVT estimator is given by 
\begin{align*}
\wthatSVT =  U \Treg V^T.
\end{align*}  
The following theorem now establishes guarantees for the singular
value thresholding estimator.

\begin{theorem}
\label{ThmSVT}
Suppose that $\pp \geq  \frac{1}{\min\{\numrows,\numcols\}}\log^7(\numrows \numcols)$. Then for any  matrix
$\wtstar \in [0,1]^{\numrows \times \numcols}$, the SVT estimator $\wthatSVT$
with threshold $\regpar = 2.1 \sqrt{\frac{\numrows+\numcols}{\pp}}$
satisfies the bound
\begin{align}
  \label{EqnSVTUpper}
\frac{1}{\numrows \numcols} \frobnorm{\wthatSVT - \wtstar}^2 \leq \UUP \min_{\wt \in [0,1]^{\numrows \times \numcols} }  \Big(
\min\Big\{ \frac{ \fnpermrank{\wt} }{\sqrt{\min \{\numrows, \numcols \} \pp} } , \frac{\fnnnrank{\wt}  }{ \min \{\numrows, \numcols \} \pp }  \Big\} +  \frac{1}{\numrows \numcols} \frobnorm{\wtstar -\wt}^2 \Big), 
\end{align}
with probability at least $1 - e^{- \UHP  \max\{\numrows,\numcols\}}$.
\end{theorem}
Observe that the bound~\eqref{EqnSVTUpper} on the risk of the SVT
estimator has the term $\sqrt{\min\{\numrows,\numcols\}}$ in the
denominator of the first expression in the minimum, as opposed to the
$\min\{\numrows,\numcols\}$ in the upper bound~\eqref{EqnOracleUpper}
from Theorem~\ref{ThmStatOracle}.  This form of
``$\sqrt{n}$-suboptimality'' arises in several permutation-based
problems of this type studied in recent papers
(e.g.,~\cite{shah2015stochastically, shah2016feeling,
  chatterjee2016estimation, shah2016permutation,
  flammarion2016optimal, pananjady2016linear}).  In some cases, this
gap---between the performance of any polynomial-time algorithm and the
best algorithm---is known to be unavoidable~\cite{shah2016feeling}
conditional on the planted clique conjecture.  It is interesting to
speculate whether such a computational complexity gap exists in the
context of the permutation-rank model.

The proof techniques underlying Theorem~\ref{ThmSVT} can
also be used to establish previously known
guarantees~\cite{koltchinskii2011nuclear,chatterjee2014matrix} for the
non-negative rank model. In order to contrast with the
permutation-rank model, let us state one such guarantee here. It is known from previous results~\cite{koltchinskii2011nuclear,chatterjee2014matrix} for the non-negative rank model that for any matrix $\wtstar \in [0,1]^{\numrows \times \numcols}$, the SVT estimator incurs an error upper bounded by
\begin{align}
  \label{EqnSVTNonnegUpper}
  \frac{1}{\numrows \numcols} \frobnorm{\wthatSVT - \wtstar}^2 \leq
  \UUP' \frac{\fnnnrank{\wtstar}}{ \min \{\numrows, \numcols \} \pp },
\end{align}
with high probability. 
On the other hand, our permutation-based modeling approach yields a stronger guarantee for the classical SVT estimator---namely, setting $\wt = \wtstar$ in our upper bound~\eqref{EqnSVTNonnegUpper} of Theorem~\ref{ThmSVT} guarantees that the
SVT estimator $\wthatSVT$ with threshold $\regpar = 2.1
\sqrt{\frac{\numrows+\numcols}{\pp}}$ satisfies the upper bound
\begin{align}
  \label{EqnCombinedBound}
  \frac{1}{\numrows \numcols} \frobnorm{\wthatSVT - \wtstar}^2 & \leq
  \UUP \min \Biggr \{ \frac{\fnnnrank{\wtstar}}{ \min \{\numrows,
    \numcols \} \pp}, \frac{\fnpermrank{\wtstar}}{ \sqrt{\min
      \{\numrows, \numcols \} \pp } } \Biggr \},
\end{align}
with high probability. The bound~\eqref{EqnCombinedBound} can yield results that are much sharper as compared to what may be obtained from the previously known guarantees~\eqref{EqnSVTNonnegUpper} for the SVT estimator. For example, suppose $\numrows = \numcols$ and $\pp = 1$. Consider the matrix $\wtstar \in [0,1]^{\numrows \times \numcols}$ given by
\begin{align*}
\wtstar_{ij} = 
\begin{cases}
1 & \quad \text{if $i > j$} \\
0 & \quad \text{if $i < j$} \\
\frac{1}{2} & \quad \text{if $i=j$}.
\end{cases}
\end{align*}
Then we have $\fnnnrank{\wtstar} = \numrows$ and $\fnpermrank{\wtstar} = 1$. Consequently, the bound~\eqref{EqnSVTNonnegUpper} from past literature yields an upper bound of
\begin{align*}
\frac{1}{\numrows \numcols} \frobnorm{\wthatSVT - \wtstar}^2 \leq \UUP',
\end{align*}
whereas in contrast, our analysis~\eqref{EqnCombinedBound} yields the sharper bound
\begin{align}
\label{EqnSVTTriMx}
\frac{1}{\numrows \numcols} \frobnorm{\wthatSVT - \wtstar}^2 \leq  \frac{\UUP \log^{2.01} \numrows}{\sqrt{\numrows}},
\end{align}
with high probability. Moreover, in our earlier work~\cite{shah2015stochastically}, we have shown that for this choice of $\wtstar$, the bound~\eqref{EqnSVTTriMx} is the best possible up to logarithmic factors for the SVT estimator with any fixed threshold $\regpar$.


\section{Properties of permutation-rank models}
\label{SecPropPermModel}

In the previous section, we established some motivating properties of permutation-based models from the perspective of statistical estimation, in this section, we derive some more insights on the
permutation-rank model.


\subsection{Comparing permutation-rank and non-negative-rank}

We begin by comparing the permutation-rank model with the conventional
non-negative rank model. To this end, first observe that the
definitions of the two models immediately imply that the
permutation-rank of any matrix is always upper bounded by its
non-negative rank, that is, for any matrix $\wt$, we have
  $\fnpermrank{\wt} \leq \fnnnrank{\wt}$. 
A natural question that now arises is whether there is any additional general condition beyond this
simple relation that
constrains the two notions of the matrix rank. The following
proposition shows that there is no other guaranteed relation between
the two notions of matrix rank.

\begin{proposition}
  \label{PropAnyRank}
  For any values $0 < \permrank \leq \nnrank \leq
  \min\{\numrows,\numcols\}$, there exist matrices whose
  permutation-rank is $\permrank$ and non-negative rank is $\nnrank$.
\end{proposition}

A particular instance that underlies part of the proof of
Proposition~\ref{PropAnyRank}, associated to any pair of values $(\permrank,\nnrank)$, is the following block matrix
$\wtmatrix_{\permrank, \nnrank}$ of size $(\numrows \times \numcols)$:
\begin{align*}
  \wtmatrix_{\permrank, \nnrank} \defn 
  \begin{bmatrix}
    \upperonesmx_{\nnrank - \permrank + 1} & 0 &
    0\\ 0 & \identitymx_{\permrank - 1} & 0\\ 0 &
    0 & 0
  \end{bmatrix},
\end{align*}
where for any value $k$,  
$\upperonesmx_{k}$ denotes an upper triangular matrix of size $(k
\times k)$ with all entries on and above the diagonal set as $1$, and
let $\identitymx_{k}$ denote the identity matrix of size $(k \times
k)$. By construction, the matrix $\wtmatrix_{\permrank, \nnrank}$ has a non-negative rank $\fnnnrank{\wtmatrix_{\permrank, \nnrank}} = \nnrank$ and a permutation rank
$\fnpermrank{\wtmatrix_{\permrank, \nnrank}} = \permrank$.

We now investigate a second relation between the two models. Recall
from our discussion earlier that the assumptions of the
permutation-rank model are much less restrictive than the assumptions
of the non-negative rank model. With this context, a natural question
that arises is to quantify the bias of an estimator that fits a matrix
of non-negative rank of $\temprank$ when the true underlying matrix
instead has a permutation rank of $\temprank$. We answer this question using the notion of the Hausdorff distance: For any two sets $\genclass_1, \genclass_2 \in \reals^{\numrows \times \numcols}$, the Hausdorff distance $\hausDis(\genclass_1,\genclass_2)$ between the two sets in the squared Frobenius norm is defined as
\begin{align}
\label{EqnDefnHausdorff}
\hausDis(\genclass_1, \genclass_2) \defn \max \Big\{ \sup \limits_{\wt
	\in \genclass_1} \ \inf \limits_{\wt' \in \genclass_2} \frobnorm{\wt
	- \wt'}^2 \ , \ \sup \limits_{\wt' \in \genclass_2} \ \inf
\limits_{\wt \in \genclass_1} & \frobnorm{\wt - \wt'}^2 \Big\}
\end{align}

The following proposition quantifies the Hausdorff distance between non-negative-rank and permutation-rank models.
\begin{proposition}
  \label{PropInclusion}
  For any positive integer $k \leq \frac{1}{2}\min\{\numcols,
  \numrows\}$, the Hausdorff distance between the sets $\nnset(k)$ and $\permset(k)$ is lower bounded as
  \begin{align}
  \label{EqnHausdorff}
 \hausDis( \nnset(k), \permset(k)) \geq \UMODEL \frac{\numrows \numcols}{k}.
  \end{align}
\end{proposition}


Proposition~\ref{PropInclusion} helps
quantify the bias on fitting a non-negative rank model when the true
matrix follows the permutation-rank model as follows. Consider any
positive integer $\temprank \leq \frac{1}{2}\min\{\numcols, \numrows\}$, and
any estimator $\wttil_\temprank$ that outputs a matrix in $\nnset(\temprank)$. Then since $\nnset(\temprank) \subseteq \permrank(\temprank)$, the
error incurred by this estimator when the true matrix lies in the set
$\permset(k)$ is lower bounded as
\begin{align}
\label{EqnInclusion}
\sup_{\wtstar \in \permset(\temprank)} \ \frac{1}{ \numrows \numcols}
\frobnorm{\wt - \wttil_{\temprank}}^2 \geq  \frac{1}{\numrows \numcols} \hausDis( \nnset(k), \permset(k)) \geq \UMODEL \frac{1}{\temprank},
\end{align}
with probability $1$.

Observe that when $k$ is a constant (but $\numrows$ and $\numcols$ are
allowed to grow), the right hand side of the
bound~\eqref{EqnHausdorff} becomes a constant, and this is the largest possible order-wise gap between any pair of matrices in
$[0,1]^{\numrows \times \numcols}$. Likewise, the right hand side of~\eqref{EqnInclusion} becomes a constant, and this is the largest possible order-wise error for any estimator that outputs matrices in $[0,1]^{\numrows \times \numcols}$.


\subsection{No ``good'' convex approximation}

In this section, we investigate a question about an important property
of the permutation-based set, and in particular, its primitive
$\permone$. There are various estimators including our regularized least squares estimator~\eqref{EqnDefnLSEPartialLowRank} as well as those studied in the literature~\cite{shah2015stochastically, shah2016feeling, shah2016permutation} which require solving a an optimization problem over the set
$\permone$. With this goal in mind, a natural question that arises is: Is the set
$\permone$ is convex? If not, then does it at least have a ``good''
convex approximation? The following proposition answers these
questions in the negative using the notion of the Hausdorff distance between sets~\eqref{EqnDefnHausdorff}.
\begin{proposition}
\label{PropNoConvex}
The Hausdorff distance~\eqref{EqnDefnHausdorff} between the set of matrices with permutation-rank one and any arbitrary convex set
$\convexclass \subseteq \reals^{\numrows \times \numrows}$ is lower bounded as
\begin{align*}
\frac{1}{\numrows \numcols} 
\hausDis ( \permone , \convexclass )
\geq
\plaincon,
\end{align*}
where $\plaincon>0$ is a universal constant. 
\end{proposition}
A specific example of a convex set $\convexclass$ is the convex hull
of $\permone$. Then by definition we have the relation $\sup
\limits_{\wt_1 \in \permone} \ \inf \limits_{\wt_2 \in \convexclass}
\frobnorm{\wt_1 - \wt_2}^2 = 0$. Consequently,
Proposition~\ref{PropNoConvex} implies that
\begin{align*}
  \sup \limits_{\wt_2 \in \convexclass} \ \inf \limits_{\wt_1 \in
    \permone} \frobnorm{\wt_1 - \wt_2}^2 = \Theta(\numrows \numcols),
\end{align*}
thus showing that the convex hull of $\permone$ is a much larger set
than $\permone$ itself.

The proof of Proposition~\ref{PropNoConvex} relies on a more general
result that we derive, one which relates a certain notion of inherent
(lack of) convexity of a set to the Hausdorff distance between that
set and any convex approximation. Note that this result does not
preclude the possibility that an optimization procedure over a convex
approximation to $\permone$ converges close enough to some element of
$\permone$ itself. We leave the investigation of this possibility to
future work.


\subsection{On the uniqueness of decomposition}

In this section, we investigate conditions for the uniqueness of the
decomposition of any matrix into its constituent components that have
a permutation-rank of one.  In the conventional setting of low
non-negative rank matrix completion, several past works~\cite{donoho2003does, theis2005first, laurberg2008theorems,
	gillis2012sparse, arora2012computing} investigate the conditions required for uniqueness of the
decomposition of matrices into their constituent non-negative rank-one
matrices.  Here we consider an
analogous question in the setting of permutation rank.  More
precisely, consider any matrix $\wt \in [0,1]^{\numrows \times
  \numcols}$ with a permutation-rank decomposition of the form
\begin{align}
\label{EqnUniqueNotation}
\wt = \sum_{\ell =1}^{\fnpermrank{\wt}} \wt^{(\ell)},
\end{align}
where $\wt^{(\ell)} \in \permset(1)$ for every $\ell \in
[\fnpermrank{\wt}]$. Under what conditions on the matrix $\wt$ is the
set $\{\wt^{(1)}, \ldots, \wt^{(\fnpermrank{\wt})} \}$ of constituent
matrices unique?  The following result provides a necessary condition
for uniqueness. In order to state the result, we use the notation $\mathbf{1}$ to denote the indicator function, that is, $\indicator{x} = 1$ if $x$ is true and $\indicator{x} = 0$ if $x$ is false.
\begin{proposition}
  \label{PropUniqueDecomposition}
A necessary condition for the uniqueness of a permutation-rank
decomposition~\eqref{EqnUniqueNotation} for any matrix $\wt$ is that for every coordinate
$(i,j) \in [\numrows] \times [\numcols]$, there is at most one $\ell
\in [\fnpermrank(\wt)]$ such that $\wt^{(\ell)}_{ij}$ is non-zero and
distinct from all other entries of $\wt^{(\ell)}$, that is,
\begin{align*}
	\sum_{\ell \in [\fnpermrank{\wt}] } \mathbf{1} \Big\{ \wt_{ij}^{\ell} \notin \{0\} \cup \{ \wt_{i'j'}^{\ell} \}_{\substack{i' \in [\numrows], j' \in [\numcols], \\ (i',j') \neq (i,j) }} \Big\} \leq 1 \qquad\quad \mbox{for every $(i,j) \in [\numrows] \times [\numcols]$}.
\end{align*}
\end{proposition}
\noindent We note that the necessary condition continues to hold even
if we restrict attention to only symmetric matrices. The necessary condition provided by 
Proposition~\ref{PropUniqueDecomposition} indicates that \emph{any}
sufficient condition for uniqueness of the decomposition must be quite strong. Moreover, we believe that the conditions for
sufficiency may be significantly stronger than those necessitated by
Proposition~\ref{PropUniqueDecomposition}. The reason for such drastic requirements for
uniqueness is the high-degree of flexibility offered by the
permutation-rank model.

Let us illustrate the necessary condition from
Proposition~\ref{PropUniqueDecomposition} with a simple
example. Consider the following matrix $\wt$ with $\numrows = \numcols
= 2$ and $\fnpermrank{\wt}=2$ and decomposition into $\wt^{(1)},
\wt^{(2)} \in \permone$:
\begin{align*}
\wt \defn
\begin{bmatrix}
1 & .6 \\ .6 & 1
\end{bmatrix}
=
\begin{bmatrix}
0 & .3 \\ .3 & .9
\end{bmatrix}
+
\begin{bmatrix}
1 & .3 \\ .3 & .1
\end{bmatrix}
\end{align*}
Observe that the necessary condition obtained in Proposition~\ref{PropUniqueDecomposition} is required to hold for every coordinate of the matrix. Let us
first evaluate this condition for the coordinate $(1,1)$ of the matrices. Since
$\wt^{(1)}_{11} = 0$, there is at most one $\ell \in \{1,2\}$ such
that $\wt^{(\ell)}_{11}$ is non-zero, and hence the coordinate $(1,1)$
satisfies the necessary condition. Moving on to coordinate $(1,2)$, we
have $\wt^{(1)}_{12} = \wt^{(1)}_{21}$ and $\wt^{(2)}_{12} = \wt^{(2)}_{21}$; hence the coordinate $(1,2)$ also
passes the necessary condition. The argument for coordinate $(1,2)$ also applies to
coordinate $(2,1)$ since the matrices involved are symmetric. We
finally test coordinate $(2,2)$. Observe that $\wt^{(1)}_{22} \notin
\{0,\wt^{(1)}_{11},\wt^{(1)}_{12},\wt^{(1)}_{21}\}$ and
$\wt^{(2)}_{22} \notin
\{0,\wt^{(2)}_{11},\wt^{(2)}_{12},\wt^{(2)}_{21}\}$. As a consequence,
for both $\ell = 1$ and $\ell = 2$, we have that $\wt^{(\ell)}_{22}$
is non-zero and distinct from all other entries of $\wt^{(\ell)}$. The
condition necessary for uniqueness is thus violated. Indeed, as
guaranteed by Proposition~\ref{PropUniqueDecomposition}, there exist
other decompositions of $\wt$ with permutation-rank 2---for instance, the decomposition
\begin{align*}
\wt =
\begin{bmatrix}
1 & .6 \\ .6 & 1
\end{bmatrix}
=
\begin{bmatrix}
0 & .4 \\
.4 & .9
\end{bmatrix}
+
\begin{bmatrix}
1 & .2 \\ .2 & .1
\end{bmatrix}
\end{align*}
is another example.

Finally, we put the negative result on the decomposition into some practical perspective with an analogy to tensor decompositions. The canonical polyadic (CP) decomposition of a tensor~\cite{hitchcock1927expression} is not unique unless strong non-degeneracy conditions are imposed~\cite{kruskal1977three}. From a theoretical perspective in many applications (for instance, in estimating latent variable models \cite{hsu2013learning}) the CP decomposition is most useful or interpretable when it is unique. Furthermore, even when the decomposition is unique, computing it is NP-hard in the worst-case~\cite{haastad1990tensor}. However, in practice the CP decomposition is often computed via ad-hoc methods that generate useful results~\cite{kolda2009tensor}.


\section{Proofs}
\label{SecProofs}

We now turn to the proofs of our main results.  In all our proofs, we
assume that the values of $\numrows$ and $\numcols$ are larger than
certain universal constants, so as to avoid subcases having to do with
small values of $(\numrows, \numcols)$. We will also ignore floors and ceilings wherever they are not critical. These assumptions entail no
ultimate loss of generality, since our results continue to hold for
all values with different constant prefactors. Throughout these and other proofs, we use the
notation $\{c, c', c_0, c_1, C, C' \}$ and so on to denote positive
constants whose values may change from line to line. 


\subsection{Proof of Theorem~\ref{ThmStatOracle}(a)}
\label{SecProofThmStatOracle}

The proof of this theorem involves generalizing an argument used in
our past work~\cite[Theorem 1]{shah2015stochastically}. In particular,
the problem setting of our past work~\cite[Theorem
  1]{shah2015stochastically} is a special case of the present problem,
restricted to the case of square matrices ($\numrows = \numcols$) and
permutation rank $\permrank = 1$.  Here we develop a number of
additional techniques in order to handle the generalization to
non-square matrices and arbitrary permutation rank.

We may assume without loss of generality that $\numrows \leq
\numcols$; otherwise, we can apply the same argument to the matrix
transposes.  It is straightforward to verify that the observation
matrix $\obs'$ can equivalently be written in the linearized form
\begin{subequations}
  \begin{align}
    \label{EqnDefnObsPrimeLowRank}
    \obs' = \wtstar + \frac{1}{\pp} \noise',
  \end{align}
where $\noise'$ has entries that are independent, and are distributed
as
\begin{align}
\label{EqnDefnWprimePartialLowRank}
[\noise']_{ij} = \begin{cases} \pp( \half - [\wtstar]_{ij}) +
  \frac{1}{2} & \qquad \mbox{with probability } \pp [\wtstar]_{ij}\\
      \pp (\half - [\wtstar]_{ij}) - \frac{1}{2} & \qquad \mbox{with
	probability } \pp (1-[\wtstar]_{ij})\\ 
      \pp (\half - [\wtstar]_{ij}) & \qquad \mbox{with probability } 1-\pp .
    \end{cases}
  \end{align}
\end{subequations}

We begin by introducing some additional notation in order to
accommodate the arbitrary permutation-rank of $\wtstar$ and the fact
that each constituent component in $\permone$ can have any arbitrary
permutation. For any pair of permutations $\pi:[\numrows] \rightarrow
[\numrows]$ and $\sig: [\numcols] \rightarrow [\numcols]$, we first
define the set
\begin{align*}
\permset(1; \pi,\sig) \defn \{ \wtmatrix \in \permone \mid \mbox{ rows
  and columns of $\wtmatrix$ are ordered according to $\pi$ and $\sig$
  respectively} \}.
\end{align*}
Now let $\piall$ denote the set of all possible permutations of
$\numcols$ items, and let $\sigall$ denote the set of all possible
permutations of the $\numrows$ users. Consider any value $\temprank
\in [\numrows]$, any sequence $\piset[\temprank] \defn
(\pi_1,\ldots,\pi_\temprank) \in \piall^\temprank$ and any sequence
$\sigset[\temprank] \defn (\sig_1,\ldots,\sig_\temprank) \in
\sigall^\temprank$. Define the set
\begin{align*}
\permset(\temprank; \piset[\temprank],\sigset[\temprank]) \defn \Big\{
\wtmatrix = \sum_{\ell = 1}^{\temprank} \wt^{(\ell)} \Big| & \mbox{
  $\wt^{(\ell)} \in \permset(1; \pi_\ell,\sig_\ell)$ for every $\ell
  \in [\temprank]$} \Big\},
\end{align*}
and an associated estimator
\begin{align*}
\wtmatrix_{\piset[\temprank],\sigset[\temprank]} \in \argmin
\limits_{\wtmatrix \in \permset(\temprank;
  \piset[\temprank],\sigset[\temprank])} \frobnorm{\obs' -
  \wtmatrix}^2.
\end{align*}  
Also define a matrix $\wtclosest$ as
\begin{align*}
\wtclosest \in \argmin_{\wt \in [0,1]^{\numrows \times \numcols}}
\Big \{ \frobnorm{\wt - \wtstar}^2 +\reg{\fnpermrank{\wt}} \Big \},
\end{align*}
as well as an associated set $\pisigset$ as
\begin{align*}
\pisigset \defn \! \Big\{\! (\temprank, \piset[\temprank],
\sigset[\temprank]) \in [\numrows] \!\times\! \piall^\temprank \!\times\!
\sigall^\temprank  \Big\vert 	 \frobnorm{\obs'\! -\! \wtmatrix_{\piset[\temprank],\sigset[\temprank]} }^2 + \reg{\temprank}\! \leq\! \frobnorm{\obs' \!- \!\wtclosest }^2 + \reg{\fnpermrank{\wtclosest}}\Big\}.
\end{align*}
Note that the set $\pisigset$ is guaranteed to be non-empty since the
parameter and permutations corresponding to $\wtclosest$ always lie in
$\pisigset$.  We claim that for any $(\temprank, \piset[\temprank],
\sigset[\temprank]) \in \pisigset$, we have
\begin{align} 
\label{EqnPartialPermReqLowRank}
\mprob \Big(
\frobnorm{\wtmatrix_{\piset[\temprank],\sigset[\temprank]} -
  \wtclosest}^2 \leq \UUP \reg{\fnpermrank{\wtclosest}} \Big) \geq 1 -
e^{- 4 \temprank \numcols \log \numcols},
\end{align}
for some positive universal constant $\UUP$.  Taking this result as
given for the moment, under our assumption of $\numcols \geq
\numrows$, for any value of $\temprank$ the cardinality of the set
$\pisigset$ restricted to any $\temprank$ is at most $e^{2 \temprank
  \numcols \log \numcols}$. Hence a union bound over all $\temprank
\in [\numrows]$ and all permutations---applied to
equation~\eqref{EqnPartialPermReqLowRank}---yields
\begin{align}
\label{EqnTowardsLSE}
\mprob \Big( \max_{(\temprank,\piset[\temprank], \sigset[\temprank])
  \in \pisigset}
\frobnorm{\wtmatrix_{\piset[\temprank],\sigset[\temprank]} -
  \wtclosest}^2 \leq \UUP \reg{\fnpermrank{\wtclosest}} \Big) \geq 1 -
e^{- \numcols \log \numcols}.
\end{align}
%
%
From the definition of the regularized least squares estimator $\wtLSE$ in equation~\eqref{EqnDefnLSEPartialLowRank} and the definition of set $\pisigset$ above, we have that $\wtLSE$ must equal
$\wtmatrix_{\piset[\temprank],\sigset[\temprank]}$ for some $(\temprank
, \piset[\temprank], \sigset[\temprank]) \in
\pisigset$. As a consequence, the tail bound~\eqref{EqnTowardsLSE} ensures that
\begin{align*}
\mprob \Big( \frobnorm{\wtLSE - \wtclosest}^2 \leq \UUP
\reg{\fnpermrank{\wtclosest}} \Big) \geq 1 - e^{- \numcols \log
  \numcols}.
\end{align*}
Finally, applying the triangle inequality yields the claimed result
\begin{align*}
\mprob \Big( \frobnorm{\wtLSE - \wtstar}^2 \leq 2\frobnorm{\wtstar -
  \wtclosest}^2 + {2\UUP \reg{\fnpermrank{\wtclosest}}} \Big) \geq 1 -
e^{- \numcols \log \numcols}.
\end{align*}


\paragraph{Proof of the bound~\eqref{EqnPartialPermReqLowRank}:}

The remainder of our proof is devoted to proving the
claim~\eqref{EqnPartialPermReqLowRank}. By definition, any
$(\temprank, \piset[\temprank], \sigset[\temprank]) \in \pisigset$
must satisfy the inequality
\begin{align*}
\frobnorm{ \obs - \wtmatrix_{\piset[\temprank],\sigset[\temprank]}}^2
+ \reg{\temprank} & \leq \frobnorm{ \obs - \wtclosest }^2 +
\reg{\fnpermrank{\wtclosest}}.
\end{align*}
Denoting the error in the estimate as
$\DelHat_{\piset[\temprank],\sigset[\temprank]} \defn
\wtmatrix_{\piset[\temprank],\sigset[\temprank]} - \wtclosest$, and
using the linearized form~\eqref{EqnDefnObsPrimeLowRank}, some
algebraic manipulations yield the basic inequality
\begin{align}
\label{EqnBasicPartialLowRank}
\frac{1}{2} \frobnorm{\DelHat_{\piset[\temprank],\sigset[\temprank]}}^2 & \leq \frac{1}{\pp}
\tracer{\noise'}{\DelHat_{\piset[\temprank],\sigset[\temprank]}} + \half \reg{(\fnpermrank{\wtclosest} - \temprank)}.
\end{align}
Now consider the set of matrices
\begin{align}
\label{eq:EqnDefnChattDiffPartialLowRank}
\diffset(\piset[\temprank],\sigset[\temprank]; \wtclosest) \defn \Big \{ \alpha(\wtmatrix - \wtclosest) \mid & \ \wtmatrix \in \permset(\temprank; \piset[\temprank], \sigset[\temprank]), \ \alpha \in [0,1] \Big \},
\end{align}
and note that $\diffset(\piset[\temprank],\sigset[\temprank]; \wtclosest) \subseteq [-1,1]^{\numrows \times   \numcols}$. For each choice of
radius $t > 0$, define the random variable
\begin{align}
\label{EqnDefnZPartialLowRank}
Z_{\piset[\temprank],\sigset[\temprank]}(t) & \defn \sup_{ \substack{\diffmx \in
		\diffset(\piset[\temprank],\sigset[\temprank]; \wtclosest),\\ \frobnorm{\diffmx} \leq t}} ~~
\frac{1}{\pp} \tracer{\noise'}{\diffmx}.
\end{align}
Using the basic inequality~\eqref{EqnBasicPartialLowRank}, the Frobenius norm error $\frobnorm{\DelHat_{\piset[\temprank],\sigset[\temprank]}}$ then satisfies the bound
\begin{align}
\label{EqnDefnDelcritPartialLowRank}
\frac{1}{2} \frobnorm{\DelHat_{\piset[\temprank],\sigset[\temprank]}}^2 & 
\leq \; Z_{\piset,\sigset} \big( \frobnorm{\DelHat_{\piset[\temprank],\sigset[\temprank]}} \big)  + \half \reg{(\fnpermrank{\wtclosest} - \temprank)}.
\end{align}
Thus, in order to obtain our desired bound, we need to
understand the behavior of the random quantity $Z_{\piset[\temprank],\sigset[\temprank]}(t)$.

Dy definition, the set $\diffset(\piset[\temprank],\sigset[\temprank]; \wtclosest)$ is ``star-shaped'',
meaning that $\alpha \diffmx \in \diffset(\piset[\temprank],\sigset[\temprank])$ for every
$\alpha \in [0,1]$ and every $\diffmx \in
\diffset(\piset[\temprank],\sigset[\temprank]; \wtclosest)$. Using this star-shaped property, we are guaranteed that $\Exs[Z_{\piset[\temprank],\sigset[\temprank]}(\delta)]$ grows at most linearly with $\delta$. We are then in turn guaranteed the existence of some scalar $\delcrit > 0$ satisfying the
critical inequality
\begin{align}
\label{EqnCriticalPartialLowRank}
\Exs[Z_{\piset[\temprank],\sigset[\temprank]}(\delcrit)] & \leq \frac{\delcrit^2}{2}.
\end{align}
Our interest is in an upper bound to the smallest (strictly) positive
solution $\delcrit$ to the critical
inequality~\eqref{EqnCriticalPartialLowRank}, and moreover, our goal is to
show that for every $t \geq \delcrit$, we have $\frobnorm{\DelHat}
\leq c \sqrt{t \delcrit}$ with high probability. To this end, define a ``bad'' event $\AuxEvent$ as
\begin{align}
\label{EqnDefnBadeventPartialLowRank}
\AuxEvent & = \big \{ \exists \Delta \in \diffset(\piset[\temprank],\sigset[\temprank]; \wtclosest) \mid
\frobnorm{\Delta} \geq \sqrt{t \delcrit} \quad \mbox{and} \quad
\frac{1}{\pp} \tracer{\noise'}{\Delta} \geq 2 \frobnorm{\Delta} \sqrt{t
	\delcrit} \big \}.
\end{align}
Using the star-shaped property of $\diffset(\piset[\temprank],\sigset[\temprank]; \wtclosest)$, it follows by a
rescaling argument that
\begin{align*}
\mprob[\AuxEvent] \leq \mprob[Z_{\piset[\temprank],\sigset[\temprank]}(\delcrit) \geq 2 \delcrit
\sqrt{t \delcrit}] \qquad \mbox{for all $t \geq \delcrit$.}
\end{align*}
The following lemma helps control the behavior of the random variable
$Z_{\piset[\temprank],\sigset[\temprank]}(\delcrit)$.
\begin{lemma}
	\label{LemZpermPartialLowRank}
	For any $\delta >0$, the mean of $Z_{\piset[\temprank],\sigset[\temprank]}(\delta)$ is bounded as
	\begin{align*}
	\Exs [ Z_{\piset[\temprank],\sigset[\temprank]}(\delta) ] \leq \plaincon_1 \frac{\max\{\temprank,\fnpermrank{\wtclosest}\} \numcols}{\pp} \log^2
	\numcols,
	\end{align*}
	and for every $u>0$, its tail probability is bounded as
	\begin{align*}
	\mprob \Big(Z_{\piset[\temprank],\sigset[\temprank]}(\delta) > \Exs[Z_{\piset[\temprank],\sigset[\temprank]}(\delta)] + u \Big) \leq \exp\Big( \frac{-\plaincon_2 u^2 \pp}{\delta^2 + \Exs[Z_{\piset[\temprank],\sigset[\temprank]}(\delta)] + u} \Big),
	\end{align*}
	where $\plaincon_1$ and $\plaincon_2$ are positive universal constants.
\end{lemma}

From this lemma, we have the tail bound
\begin{align*}
\mprob \Big(Z_{\piset[\temprank],\sigset[\temprank]}(\delcrit) > \Exs[Z_{\piset[\temprank],\sigset[\temprank]}(\delcrit)] + \delcrit
\sqrt{t \delcrit} \Big) \leq \exp\Big( \frac{-\plaincon_2 (\delcrit \sqrt{t \delcrit})^2 \pp}{ \delcrit^2 + \Exs[Z_{\piset[\temprank],\sigset[\temprank]}(\delcrit)] +  (\delcrit \sqrt{t \delcrit})} \Big), \quad \mbox{for all $t > 0$.}
\end{align*}
By the definition of $\delcrit$ in~\eqref{EqnCriticalPartialLowRank}, we have
$\Exs[Z_{\piset[\temprank],\sigset[\temprank]}(\delcrit)] \leq \delcrit^2 \leq \delcrit \sqrt{t \delcrit}$
for all $t \geq \delcrit$, and consequently
\begin{align}
\label{EqnAuxEventBound}
\mprob[\AuxEvent] & \leq \mprob[Z_{\piset[\temprank],\sigset[\temprank]}(\delcrit) \geq 2 \delcrit \sqrt{t
	\delcrit} \big] \; \leq \; \exp\Big( \frac{-\plaincon_2 (\delcrit \sqrt{t \delcrit})^2 \pp }{3\delcrit \sqrt{t \delcrit} } \Big), \quad \mbox{for all $t \geq \delcrit$.}
\end{align}
Now we must have either $\frobnorm{\DelHat_{\piset[\temprank],\sigset[\temprank]}} \leq \sqrt{t
	\delcrit}$, or we have $\frobnorm{\DelHat_{\piset[\temprank],\sigset[\temprank]}} > \sqrt{t
	\delcrit}$. In the latter case, conditioning on the complement
$\AuxEvent^c$, our basic inequality~\eqref{EqnBasicPartialLowRank} implies that 
\begin{align*}
\frac{1}{2}
\frobnorm{\DelHat_{\piset[\temprank],\sigset[\temprank]}}^2 \leq 2 \frobnorm{\DelHat_{\piset[\temprank],\sigset[\temprank]}} \sqrt{t
	\delcrit} + \half \reg{(\fnpermrank{\wtclosest} - \temprank)},
\end{align*}
and hence 
\begin{align}
\label{EqnDelhatBound}
\frobnorm{\DelHat_{\piset[\temprank],\sigset[\temprank]}} \leq 4 \sqrt{t \delcrit} + \sqrt{ \reg{(\fnpermrank{\wtclosest} - \temprank)} }.
\end{align}  
Putting together the bounds~\eqref{EqnAuxEventBound} and~\eqref{EqnDelhatBound} then yields
\begin{align}
\mprob \Big( \frobnorm{\DelHat_{\piset[\temprank],\sigset[\temprank]}}^2 \leq 32 t \delcrit +  2 \reg{(\fnpermrank{\wtclosest} - \temprank)} \Big)
\geq 1 - \exp \big( -\frac{\plaincon_2}{3} \delcrit \sqrt{t \delcrit} \pp \big),  \quad \mbox{for all $t \geq \delcrit$.}
\label{EqWithDelCritHPPartialLowRank}
\end{align}
Finally, from the bound on the expected value of $Z_{\piset[\temprank],\sigset[\temprank]}(t)$ in
Lemma~\ref{LemZpermPartialLowRank}, we see that the critical
inequality~\eqref{EqnCriticalPartialLowRank} is satisfied for 
\begin{align*}
\delcrit = \sqrt{\frac{\plaincon_1 \max\{\fnpermrank{\wtclosest}, \temprank\} \numcols}{\pp}}\log \numcols. 
\end{align*}
Setting $t = \plaincon' \delcrit$
in~\eqref{EqWithDelCritHPPartialLowRank} for a large enough constant $\plaincon'$ yields
\begin{align}
\mprob \Big( \frobnorm{\DelHat_{\piset[\temprank],\sigset[\temprank]}} \leq \frac{\plaincon_1'
	\fnpermrank{\wtclosest} \numcols}{\pp} \log^{2.01} \numcols \Big) \geq 1 - \exp\Big( - 4 \max\{\fnpermrank{\wtclosest}, \temprank\} \numcols \log \numcols \Big),
\end{align}
for some constant $\plaincon_1'>0$, thus proving the bound~\eqref{EqnPartialPermReqLowRank}.\\
\noindent It remains to prove Lemma~\ref{LemZpermPartialLowRank}.


\paragraph*{Proof of Lemma~\ref{LemZpermPartialLowRank}}

We break our proof into two parts, corresponding to bounds on the mean
of the random variable
$Z_{\piset[\temprank],\sigset[\temprank]}(\delta)$ followed by control
of its tail behavior.\\

\noindent \underline{Bounding the mean}: We begin by establishing an upper bound
on the mean $\Exs[ Z_{\piset[\temprank],\sigset[\temprank]}(\delta)]$. In order to obtain the desired upper bound, in the proof we carefully account for the rank $\temprank$ and the different permutations associated to any matrix in $\diffset(\piset[\temprank],\sigset[\temprank]; \wtclosest)$. 

For convenience of analysis, we introduce a new random variable 
\begin{align*}
\widetilde{Z}_{\piset[\temprank],\sigset[\temprank]} \defn \sup_{ \diffmx \in  \diffset({\piset[\temprank],\sigset[\temprank]}; \wtclosest)} \tracer{\noise'}{\diffmx}.
\end{align*} 
Then by definition, we have
$\Exs[Z_{\piset[\temprank],\sigset[\temprank]}(\delta)] \leq \frac{1}{\pp}
\Exs[\widetilde{Z}_{\piset[\temprank],\sigset[\temprank]}]$ for every $\delta>0$. In addition, since $\wtclosest \in \permset(\temprank)$, it can be decomposed as $\wtclosest = \sum_{\ell = 1}^{\temprank} \wtclosest^{(\ell)}$, for some matrices $\wtclosest^{(1)},\ldots,\wtclosest^{(\temprank)} \in \permset(1)$. 

We introduce some additional notation for ease of exposition. If $\fnpermrank{\wtclosest} < \temprank$, then let $\wtclosest^{(\fnpermrank{\wtclosest}+1)},\ldots,\wtclosest^{\temprank}$ denote all-zero matrices. Hence we can write $\wtclosest = \sum_{\ell = 1}^{\max\{\fnpermrank{\wtclosest},\temprank\} } \wtclosest^{(\ell)}$. If $\fnpermrank{\wtclosest} > \temprank$ then let $\pi_{\temprank+1},\ldots,\pi_{\fnpermrank{\wtclosest}}$ be arbitrary (but fixed) permutations of $\numrows$ items and $\sigma_{\temprank+1},\ldots,\sigma_{\fnpermrank{\wtclosest}}$ be arbitrary (but fixed) permutations of $\numcols$ items. With this notation in place, we have the following deterministic upper bound on the value of the random variable $\widetilde{Z}_{\piset[\temprank],\sigset[\temprank]}$: 
\begin{align*}
\widetilde{Z}_{\piset[\temprank],\sigset[\temprank]} \leq \sum_{\ell= 1}^{\max\{\fnpermrank{\wtclosest},\temprank\}} \sup_{ [\diffmx]_\ell \in  \diffset(\{\pi_\ell\},\{\sig_\ell\}; \wtclosest^{(\ell)})}  \tracer{\noise'}{[\diffmx]_\ell}.
\end{align*}
We also recall our assumption that $\numcols \geq \numrows$ without loss of generality. Now let $\metent(\epsilon,\mathbb{C},\frobnorm{\cdot})$
denote the $\epsilon$ metric entropy of class $\mathbb{C} \subset
\reals^{\numrows \times \numcols}$ in the Frobenius norm metric
$\frobnorm{\cdot}$.  Then the truncated form of Dudley's entropy
integral inequality yields\footnote{Here we use $(\Delta \epsilon)$ to denote the differential of $\epsilon$, so as to avoid confusion with the number of columns $\numcols$.}
\begin{align}
\label{EqnDudleyPartialLowRank}
\Exs[ \widetilde{Z}_{\piset[\temprank],\sigset[\temprank]}] & \leq \sum_{\ell = 1}^{\max\{\fnpermrank{\wtclosest},\temprank\}} \plaincon \; \Big \{ \numcols^{-8}
+ \int_{\frac{1}{2} \numcols^{-9}}^{2 \numcols} \sqrt{
	\metent(\epsilon,\diffset(\{\pi_\ell\},\{\sig_\ell\}; \wtclosest^{(\ell)}),\frobnorm{.})} (\Delta \epsilon) \Big \},
\end{align}
where we have used the fact that the diameter of the set $\diffset(\{\pi_\ell\},\{\sig_\ell\}; \wtclosest^{(\ell)})$ is at most $2 \numcols$ in the Frobenius norm.

\arxiv{In }\nips{The paper}\arxiv{our past work}~\cite[Lemma 2]{shah2015stochastically}\arxiv{, we derived}\nips{ derives} a bound on the metric entropy of the set $\diffset(\{\pi_\ell\},\{\sig_\ell\}; \wtclosest^{(\ell)})$ as:
\begin{align*}
\metent \Big( \epsilon, \diffset(\{\pi_\ell\},\{\sig_\ell\};\wtclosest^{(\ell)}), \frobnorm{\cdot} \Big) & \leq 16 \frac{\numcols^2}{
	\epsilon^2} \big(\log \frac{\numcols}{ \epsilon} \big)^2,
\end{align*}
for any $\epsilon > 0$ and $\ell \in [\temprank]$. 
Substituting this bound on the metric entropy into the Dudley bound~\eqref{EqnDudleyPartialLowRank} yields
\begin{align*}
\Exs [\widetilde{Z}_{\piset[\temprank],\sigset[\temprank]} ] & \leq \plaincon' \max\{\fnpermrank{\wtclosest},\temprank\} \numcols \log^2 \numcols.
\end{align*}
The inequality $\Exs[Z_{\piset[\temprank],\sigset[\temprank]}(\delta)]
\leq \frac{1}{\pp}
\Exs[\widetilde{Z}_{\piset[\temprank],\sigset[\temprank]}]$ then
yields the claimed result.~\\

\vspace*{.1in}

\noindent \underline{Bounding the tail}: In order to establish the
claimed tail bound on the deviations of
$Z_{\piset[\temprank],\sigset[\temprank]}(\delta)$ above its mean, we
  use a Bernstein-type bound on the supremum of empirical processes
  due to Klein and Rio~\cite[Theorem 1.1c]{klein2005concentration},
  which we state in a simplified form here.
\begin{lemma}
  \label{LemKleinRio}
  Let $X \defn (X_1,\ldots,X_m)$ be any sequence of zero-mean,
  independent random variables, each taking values in $[-1,1]$. Let
  $\mathcal{V} \subset [-1,1]^m$ be any measurable set of $m$-length
  vectors.  Then for any $u>0$, the supremum $X^\dagger = \sup_{v \in
    \mathcal{V} } \inprod{X}{v}$ satisfies the upper tail bound
	\begin{align*}
	\mprob \big( X^\dagger > \Exs[ X^\dagger] + u \big) \leq \exp
	\Big(\frac{ - u^2 }{2\sup_{v \in \mathcal{V} } \Exs[\inprod{v}{X}^2] +
		4\Exs[ X^\dagger] + 3 u} \Big).
	\end{align*}
\end{lemma}

We now call upon Lemma~\ref{LemKleinRio} setting $\mathcal{V} = \{ \diffmx \in \diffset(\piset[\temprank],\sigset[\temprank];\wtclosest) \mid \frobnorm{\diffmx} \leq \delta \}$, $X
=  \noise'$, and $X^\dagger = \pp Z_{\piset[\temprank],\sigset[\temprank]}(\delta)$. The entries of the
matrix $\noise'$ are mutually independent, have a mean of zero, and are bounded by $1$ in absolute value. Then we have
$\Exs[X^\dagger] = \pp \Exs[Z_{\piset[\temprank],\sigset[\temprank]}(\delta) ]$ and
$\Exs[\tracer{\diffmx}{\noise'}^2] \leq 4 \pp
\frobnorm{\diffmx}^2 \leq 4 \pp \delta^2$ for every $\diffmx
\in \mathcal{V}$. With these assignments, and some algebraic manipulations, we obtain that for every $u > 0$,
\begin{align*}
\mprob \big( Z_{\piset[\temprank],\sigset[\temprank]}(\delta) > \Exs[Z_{\piset[\temprank],\sigset[\temprank]}(\delta)] + u  \big)
\leq \exp \Big( \frac{ - u^2 \pp }{8  \delta^2 + 4\Exs[Z_{\piset[\temprank],\sigset[\temprank]}(\delta)]
	+ 3u} \Big),
\end{align*}
as claimed.


\subsection{Proof of Theorem~\ref{ThmStatOracle}(b)}

Assume without loss of generality that $\numcols \geq
\numrows$. Throughout the proof, we ignore floor and ceiling
conditions as these are not critical to the proof and affect the lower
bound by only a constant factor.

The Gilbert-Varshamov bound~\cite{gilbert1952comparison, varshamov1957estimate} from coding theory guarantees the existence of 
\begin{align*}
\packnum \defn \exp\Big({\plaincon (\numcols \nnrank + \pp \oracleeps^2)}\Big)
\end{align*}
binary vectors $\packvec^1,\ldots,\packvec^\packnum$, each of length $(\numcols \nnrank + \pp \oracleeps^2)$, such that the Hamming distance between any pair of vectors in this set is lower bounded as
\begin{align*}
\hamming(\packvec^\ell, \packvec^{\ell'}) \geq \frac{\numcols \nnrank + \pp \oracleeps^2}{10}.
\end{align*}
For some $\delta \in (0,\frac{1}{4})$ whose value is specified later, define a related set of vectors $\widetilde{\packvec}^1,\ldots,\widetilde{\packvec}^\packnum$ as 
\begin{align*}
\widetilde{\packvec}^\ell_j = 
\begin{cases}
\half + \delta & \quad \mbox{if } \packvec^\ell_j = 1 \\
\half - \delta & \quad \mbox{if } \packvec^\ell_j = 0,
\end{cases}
\end{align*}
for every $\ell \in [\packnum]$ and $j \in [\numcols \nnrank + \pp \oracleeps^2]$.
Next define a set of ``low rank'' matrices $\packmat^1,\ldots,\packmat^\packnum \in [0,1]^{\numrows \times \numcols}$ where the matrix $\packmat^\ell$ is obtained as follows. For each $\ell \in [\packnum]$, arrange the first $\nnrank \numcols$ entries of vector $\widetilde{\packvec}^\ell$ as the entries of an $(\nnrank \times \numcols)$ matrix---this arrangement may be done in an arbitrary manner as long as it is consistent across every $\ell \in [\packnum]$. Now append a $(\frac{\pp \oracleeps^2}{\numcols} \times \numcols)$ matrix at the bottom, whose entries comprise the last $\pp \oracleeps^2$ entries of the vector $\widetilde{\packvec}^\ell$---again, this arrangement may be done in an arbitrary manner as long as it is consistent across every $\ell \in [\packnum]$. Now stack $\frac{1}{\pp}$ copies of the resulting $\big( (\nnrank+ \frac{\pp \oracleeps^2}{\numcols} ) \times \numcols\big)$ matrix on top of each other to form a $\big((\frac{\nnrank}{\pp} + \frac{\oracleeps^2}{\numcols}) \times \numcols\big)$ matrix. Note that our assumption $\oracleeps^2 + \frac{\nnrank \max\{\numrows, \numcols\}}{\pp} \leq \numrows \numcols$, along with the assumption $\numcols \geq \numrows$, implies that $\numrows \geq \frac{\nnrank}{\pp}  +  \frac{\oracleeps^2}{\numcols}$. Append $(\numrows - (\frac{\nnrank}{\pp} + \frac{\oracleeps^2}{\numcols}))$ rows of all zeros at the bottom of this matrix, and denote the resultant $(\numrows \times \numcols)$ matrix as $\packmat^\ell$. 

We now show that $\packmat^\ell \in \frobballnn{\nnrank}{\oracleeps}$ for every $\ell \in [\packnum]$, that is, we show that the matrix $\packmat^\ell \in [0,1]^{\numrows \times \numcols}$ can be decomposed into a sum of a low-rank matrix (of non-negative rank at most $\nnrank$) and a sparse matrix (number of non-zero entries at most $\oracleeps^2$). First we set to zero the entries in $\packmat^\ell$ which correspond to the last $\pp \epsilon^2$ entries of the vector $\widetilde{\packvec}^\ell$. Let us denote the resulting matrix as $\widetilde{\packmat}^\ell$. Each row of the matrix $\widetilde{\packmat}^\ell$ is either all zero or is identical to one among the first $\nnrank$ rows of $\packmat^\ell$. Consequently we have $\fnnnrank{\widetilde{\packmat}^\ell} \leq \nnrank$. Also observe that in the matrix $(\packmat^\ell - \widetilde{\packmat}^\ell)$, the number of non-zero entries is at most $\frac{1}{\pp} \times \pp \epsilon^2 = \epsilon^2$, and furthermore, each of these entries lie in the interval $[0,1]$. Hence we have $\frobnorm{\packmat^\ell - \widetilde{\packmat}^\ell}^2 \leq \epsilon^2$. The matrix $\packmat^\ell$ thus satisfies all the requirements for membership in the set $\frobballnn{\nnrank}{\oracleeps}$.

For every $\ell \in [\packnum]$, let $\mprob^\ell$ denote the probability distribution of the matrix $\obs$ obtained by setting $\wtstar = \packmat^\ell$. One can verify that the set of matrices $\packmat^1,\ldots,\packmat^\packnum$ constructed above has the following two properties, for every pair $\ell \neq \ell' \in [\packnum]$:
\begin{align*}
\kl{\mprob^\ell}{\mprob^{\ell'}} \leq \plaincon' \delta^2 \pp \big( \frac{\numcols \nnrank}{\pp} + \oracleeps^2 \big),
\end{align*}
and
\begin{align*}
\frobnorm{\packmat^\ell - \packmat^{\ell'}}^2 \geq  \frac{\delta^2}{10} \big( \frac{\numcols \nnrank}{\pp} + \oracleeps^2 \big).
\end{align*}
Substituting these relations in Fano's
inequality~\cite{cover2012elements} yields that when $\wtstar$ is
drawn uniformly at random from the set
$\{\packmat^1,\ldots,\packmat^\packnum\}$, any estimate $\wthat$ for
$\wtstar$ has squared Frobenius error at least
\begin{align*}
\Exs[ \frobnorm{\wthat - \wtstar}^2] \geq \frac{\delta^2}{20} \Big(
\frac{\numcols \nnrank}{\pp} + \oracleeps^2 \Big) \Big( 1 - \frac{
  \plaincon' \delta^2 \pp \big( \frac{\numcols \nnrank}{\pp} +
  \oracleeps^2 \big) + \log 2}{ \plaincon (\numcols \nnrank + \pp
  \oracleeps^2) } \Big) \stackrel{(i)}{\geq} \plaincon'' \big(
\frac{\numcols \nnrank}{\pp} + \oracleeps^2 \big),
\end{align*}
where inequality $(i)$ is obtained by choosing $\delta^2$ as a small enough constant (that depends only on $\plaincon$ and $\plaincon'$). Recalling our assumption $\numcols \geq \numrows$, and consequently replacing $\numcols$ by $\max\{\numrows, \numcols\}$ in the bound yields the claimed result.

\subsection{Proof of Theorem~\ref{ThmSVT}}

We now turn to analysis of the singular-value thresholding (SVT)
estimator.  This proof is based on the framework of a proof from our
earlier work~\cite[Theorem 2]{shah2015stochastically}, which can be
seen as a particular case with $\numrows = \numcols$ and $\permrank =
1$. We introduce certain additional tricks in order to generalize the
proof for general values of $\permrank$ and to obtain a sharp
dependence on $\permrank$.  As in our previous proofs, we may assume
without loss of generality that $\numrows \leq \numcols$.

Recall from equation~\eqref{EqnDefnObsPrimeLowRank} that we can write our observation model as $\obs' = \wtstar + \frac{1}{\pp} \noise'$,
where $\noise' \in [-1,1]^{\numrows \times \numcols}$ is a zero-mean matrix
with mutually independent entries.  Also recall
that these entries follow the distribution
\begin{align}
[\noise']_{ij} = 
\begin{cases} \pp(
\half - [\wtstar]_{ij}) + \frac{1}{2} & \qquad \mbox{with probability
} \pp [\wtstar]_{ij}\\ 
\pp (\half - [\wtstar]_{ij}) - \frac{1}{2} & \qquad \mbox{with
	probability } \pp (1-[\wtstar]_{ij})\\ 
\pp (\half - [\wtstar]_{ij}) & \qquad \mbox{with probability } 1-\pp .
\end{cases}
\label{EqnDefnWprimePartialLowRankRepeat}
\end{align}
For any matrix $A$, let $\singularvalue{1}{A}, \singularvalue{2}{A},\ldots$ denote its singular values in descending order. 

Our proof of the upper bound is based on four lemmas. The first lemma
is a result from \arxiv{our earlier work}\nips{the paper}~\cite{shah2015stochastically}.
\begin{lemma}(\cite[Lemma 3]{shah2015stochastically})
  \label{LemSTSVD}
  If $\regpar \geq \frac{1.01 \opnorm{\noise'}}{\pp}$, then
  \begin{align*}
    \frobnorm{\wthatSVT - \wtstar}^2 & \leq \plaincon \sum_{j=1}^\numrows \min \big \{ \regpar^2, \sigma_j^2(\wtstar) \big \}
  \end{align*}
  with probability at least $1-c_1 e^{-c' \numrows}$, where $\plaincon$, $c_1$ and $c'$ are positive universal constants.
\end{lemma}

\noindent Our second lemma is an approximation-theoretic result that bounds the tail of the singular values of any matrix with a given permutation-rank or non-negative rank. The proof of this lemma builds on a construction due to Chatterjee~\cite{chatterjee2014matrix}.
\begin{lemma}
	\label{LemSTSVDChatterjee}
	(a) For any matrix $\wt \in \permset(\permrank)$ and any $s \in \{1, 2,
	\ldots, \numrows - 1 \}$, we have
	\begin{align*}
	\sum_{j=s+1}^\numrows \sigma^2_j(\wt) & \leq
	\frac{\numrows \numcols \permrank^2}{s}.
	\end{align*}
	(b) For any matrix $\wt \in \nnset(\nnrank)$ and any $s \in \{1, 2,
	\ldots, \numrows - 1 \}$, we have
	\begin{align*}
	\sum_{j=s+1}^\numrows \sigma^2_j(\wt) & \leq
	\numrows \numcols \max\big\{ \frac{\nnrank - s}{\nnrank}, 0\big\}.
	\end{align*}
\end{lemma}

\noindent Our third lemma controls the noise term $\noise'$.
\begin{lemma}
  \label{LemWOp}
  Suppose that $\pp \geq \frac{1}{\min\{\numrows,\numcols\}}\log^7(\numrows \numcols)$. Then given a random matrix $\noise'$ with entries distributed according
  to the distribution~\eqref{EqnDefnWprimePartialLowRankRepeat}, we
  have
  \begin{align*}
    \mprob \big( \opnorm{ \noise' } > 2.01 \sqrt{\pp (\numrows +
      \numcols) } \big) \leq e^{- \plaincon' \max\{\numrows,
      \numcols\}}.
  \end{align*}
\end{lemma}

Finally, our fourth lemma is a more general relation pertaining to matrices established in the paper~\cite{schonemann1968two}. 
\begin{lemma}[\cite{schonemann1968two}]
	\label{LemAlignedIsClosest}
	For any pair of matrices $\mxtempone, \mxtemptwo \in \reals^{\numrows \times \numcols}$ with singular value decompositions $\mxtempone = \mxU_{1} \mxdiag_{1} \mxV_{1}^T$ and $\mxtemptwo = \mxU_2 \mxdiag_2 \mxV_2^T$, it must be that
	\begin{align}
	\label{EqnAlignedIsClosest}
	(\mxU_{1} \mxU_{2}^T, \ \mxV_1 \mxV_2^T)	\in & \argmin_{ \mxU \in \reals^{\numrows \times \numrows}, \  \mxV \in \reals^{\numcols \times \numcols}  } \frobnorm{\mxtempone - \mxU \mxtemptwo \mxV^T}^2 \\ & \mbox{such that }\mxU^T = \mxU^{-1},\ \mxV^T = \mxV^{-1}. \nonumber
	\end{align}
\end{lemma}

Based on these four lemmas, we now complete the proof of the
theorem. From Lemma~\ref{LemWOp} we see that the choice $\regpar = 2.1
\sqrt{\frac{\numrows + \numcols}{\pp}}$ guarantees that $\regpar \geq
\frac{1.01\opnorm{\noise'}}{\pp}$ with probability at least $1 -
e^{-\plaincon' \max\{\numrows, \numcols\}}$. Consequently, the
condition required for an application of Lemma~\ref{LemSTSVD} is
satisfied, and applying this lemma then yields the upper bound
\begin{align}
\label{LemSVTMstarProved}
\frobnorm{\wthatSVT - \wtstar}^2 & \leq \plaincon \sum_{j=1}^\numrows
\min \big \{ \frac{\numcols}{\pp}, \sigma_j^2(\wtstar) \big \}
\end{align}
with probability at least $1 - e^{-\plaincon' \numcols}$, where we have also used our assumption that $\numrows \leq \numcols$. 

Now consider any matrix $\wtclosest \in \reals^{\numrows \times \numcols}$. In what follows, we convert the bound~\eqref{LemSVTMstarProved} into one that depends on the properties of $\wtclosest$, namely $\fnpermrank{\wtclosest}$, $\fnnnrank{\wtclosest}$ and $\frobnorm{\wtstar - \wtclosest}$. Let $\permrank_0 = \fnpermrank{\wtclosest}$ and $\nnrank_0 = \fnnnrank{\wtclosest}$.

We first have the following deterministic upper bound
\begin{align}
\sum_{j=1}^\numrows
\min \big \{ \frac{\numcols}{\pp}, \sigma_j^2(\wtstar) \big \} & \leq 2 \sum_{j=1}^\numrows
\min \big \{ \frac{\numcols}{\pp}, \sigma_j^2(\wtclosest) \big \} \nonumber \\ 
&~~~ + 2 \sum_{j=1}^\numrows
\Big( \min \big \{ \sqrt{\frac{\numcols}{\pp}}, \sigma_j(\wtstar) \big \} - \min \big \{ \sqrt{\frac{\numcols}{\pp}}, \sigma_j(\wtclosest) \big  \} \Big)^2 \nonumber \\
& \leq 2 \sum_{j=1}^\numrows
\min \big \{ \frac{\numcols}{\pp}, \sigma_j^2(\wtclosest) \big \}  + 2 \sum_{j=1}^\numrows
\big( \sigma_j(\wtstar)  - \sigma_j(\wtclosest) \big)^2,
\label{EqnSVTTriangle}
\end{align}
where the inequality~\eqref{EqnSVTTriangle} is a consequence of the more general result that $(\min\{a,b_1\} - \min\{a,b_2\})^2 \leq (b_1 - b_2)^2$ for any three real numbers $a$, $b_1$, and $b_2$.

We now bound the two terms on the right hand side of~\eqref{EqnSVTTriangle} separately. For the second term, we call upon Lemma~\ref{LemAlignedIsClosest} with the choices $\mxtempone = \wtstar$ and $\mxtemptwo = \wtclosest$. With this choice, some simple algebra yields that the minimum value of the objective in~\eqref{EqnAlignedIsClosest} equals $\sum_{j=1}^{\numrows} ( \singularvalue{j}{\wtstar} - \singularvalue{j}{\wtclosest} )^2$. On the other hand, the choice of $\mxU$ and $\mxV$ as identity matrices is feasible for~\eqref{EqnAlignedIsClosest}, and the associated value of the objective equals $\frobnorm{\wtstar -\wtclosest}^2$. Consequently, we have the inequality
\begin{align}
\sum_{j=1}^{\numrows} ( \singularvalue{j}{\wtstar} - \singularvalue{j}{\wtclosest} )^2 \leq \frobnorm{\wtstar -\wtclosest}^2.
\label{EqnSVTTriangleSecondTerm}
\end{align}
As for the first term of~\eqref{EqnSVTTriangle}, an application of Lemma~\ref{LemSTSVDChatterjee}(a) to the matrix $\wtclosest$ yields the bound 
\begin{align}
\sum_{j=1}^\numrows
\min \big \{ \frac{\numcols}{\pp}, \sigma_j^2(\wtclosest) \big \}
 & \leq  \min_{s \in
  [\numrows]} \big( \frac{s \numcols}{\pp} + \frac{(\permrank_0)^2
  \numrows \numcols}{s} \big) \nonumber \\
&  \leq 3 \permrank_0 \numcols  \sqrt{\frac{\numrows}{\pp} }, \label{EqnSVTFirstMin}
\end{align}
where inequality~\eqref{EqnSVTFirstMin} is obtained with the choice $s =
\lceil \permrank_0 \sqrt{\pp \numrows} \rceil$. 
Separately, an application of Lemma~\ref{LemSTSVDChatterjee}(b) to the matrix $\wtclosest$ yields
\begin{align}
\sum_{j=1}^\numrows
\min \big \{ \frac{\numcols}{\pp}, \sigma_j^2(\wtclosest) \big \} & \leq \min_{s \in
  [\numrows]} \big( \frac{s \numcols}{\pp} + \numrows \numcols
\max\big\{ 1 - \frac{s}{\nnrank_0}, 0\big\} \big) \nonumber \\
& \leq \frac{\nnrank_0 \numcols }{\pp }, \label{EqnSVTSecondMin}
\end{align}
where  the inequality~\eqref{EqnSVTFirstMin} is obtained with the choice $s = \nnrank_0$. Combining the bounds~\eqref{LemSVTMstarProved},~\eqref{EqnSVTTriangle},~\eqref{EqnSVTTriangleSecondTerm},~\eqref{EqnSVTFirstMin} and~\eqref{EqnSVTSecondMin}, we obtain the result that the inequality
\begin{align*}
\frobnorm{\wthatSVT - \wtstar}^2 \leq 2 \min\Big\{ \frac{3 \permrank_0 \numcols  \sqrt{\numrows} } {\sqrt{\pp} } , \frac{\nnrank_0 \numcols }{\pp  }  \Big\} + 2\frobnorm{\wtstar -\wtclosest}^2, 
\end{align*}
must hold with probability at least $1 - e^{\plaincon' \numcols}$.  Finally, recalling our assumption that $\numcols \geq \numrows$ and
substituting $\numrows = \min \{\numrows, \numcols\}$ and $\numcols =
\max \{ \numrows, \numcols \}$ yields the claimed result.


\paragraph{Proof of Lemma~\ref{LemSTSVDChatterjee}}
\underline{Part (a)}: Without loss of generality, assume that
$\numcols \geq \numrows$.

We begin with an upper bound on the tail of the singular values of any matrix in $\permset(1)$, that is, of any matrix that has a permutation-rank of $1$. The proof of this bound uses a construction due to
Chatterjee~\cite{chatterjee2014matrix} for a rank $\widetilde{s}$ approximation of any matrix in $\permset(1)$, for any value $\widetilde{s} \in [\numrows]$. We first reproduce Chatterjee's construction.

For a given matrix $\wt \in \permset(1)$, 
define the vector $\tau \in \real^\numcols$ of column sums---namely,
with entries $\tau_j = \sum_{i=1}^\numrows [\wt]_{ij}$ for $j \in
[\numcols]$.  Using this vector, define a rank $\widetilde{s}$
approximation $\widetilde{\wt}$ to $\wt$ by grouping the
columns according to the vector $\tau$ according to the following procedure:
\begin{itemize}
\item
  Observing that each $\tau_j \in [0,\numrows]$, divide the full
  interval $[0,\numrows]$ into $\widetilde{s}$ groups---say of the
  form
  \begin{align*}
    [0,\numrows/\widetilde{s}), [\numrows/\widetilde{s},
        2\numrows/\widetilde{s}), \ldots
        [(\widetilde{s}-1)\numrows/\widetilde{s},\numrows].
  \end{align*}
  If $\tau_j$ falls into the interval $\alpha$ for some $\alpha \in
  [\widetilde{s}]$, then map column $j$ to the group $G_\alpha$ of
  indices.
    \item For each $\alpha \in [\widetilde{s}]$ such that group
      $G_\alpha$ is non-empty, choose a particular column index $j' 
      \in G_\alpha$ in an \emph{arbitrary} fashion.  For every other
      column index $j \in G_\alpha$, set $\widetilde{\wt}_{ij} =
      \wt_{i j'}$ for all $i \in [\numrows]$.
\end{itemize}

\vspace*{.04in}

By construction, the matrix $\widetilde{\wtmatrix}$ has at most
$\widetilde{s}$ distinct rows, and hence rank at most $\widetilde{s}$.
Now consider any column $j \in [\numcols]$ and suppose that $j \in
G_\alpha$ for some $\alpha \in [\widetilde{s}]$. Let $j'$ denote the column chosen for the group $G_\alpha$
in the second step of the construction. Since $\wt \in \permone$, we
must either have $\wt_{ij} \geq \wt_{ij'} =
\widetilde{\wtmatrix}_{ij}$ for every $i \in [\numrows]$, or $\wt_{ij}
\leq \wt_{ij'} = \widetilde{\wt}_{ij}$ for every $i \in
     [\numrows]$. Then we are guaranteed that
\begin{align}
\label{eq:STSVDChatterjeeRankOneAbs}
\sum_{i=1}^\numrows | \widetilde{\wt}_{ij} - \wt_{ij}| = \mid
\sum_{i=1}^\numrows ( \widetilde{\wt}_{ij} - \wt_{ij} ) \mid = |
\tau_{j'} - \tau_{j}| \leq \frac{\numrows}{\widetilde{s}},
\end{align}
where we have used the fact the pair $(\tau_j, \tau_{j'})$ must lie in
an interval of length at most $\numrows/\widetilde{s}$.  This
completes the description of Chatterjee's construction.

In what follows, we use Chatterjee's result in order to obtain our
claimed bound on the tail of the spectrum of any matrix $\wt \in
\permset(\permrank)$. We modify the result in a careful 
manner that allows us to obtain the desired dependence on the
parameter $\permrank$. Recall that any matrix $\wt \in
\permset(\permrank)$ can be decomposed as
\begin{align*}
\wtmatrix = \sum_{\ell=1}^{\permrank} \wt^{(\ell)},
\end{align*}
for some matrices $\wt^{(1)},\ldots,\wt^{(\permrank)} \in \permset(1)$. Let $\widetilde{s} = \frac{s}{\permrank}$. For every $\ell \in [\nnrank]$, let $\widetilde{\wt}^{(\ell)}$ be a rank $\widetilde{s} = \frac{s}{\permrank}$ approximation of $\wt^{(\ell)}$ obtained from Chatterjee's construction above, but with the following additional detail. Observe that in Chatterjee's construction, the choice of column $j'$ from group $G_\alpha$ is arbitrary. For our construction, we will make a specific choice of this column: we choose the column whose entries have the smallest values among all columns in the group $G_\alpha$. With this choice, we have the property 
\begin{align}
\label{EqnMonotoneApprox}
\widetilde{\wt}^{(\ell)}_{ij} \leq \wtmatrix^{(\ell)}_{ij} \qquad \mbox{for every $\ell \in [\permrank],\ i \in [\numrows], j \in [\numcols]$}.
\end{align}

Now let $\widetilde{\wtmatrix} \defn \sum_{\ell=1}^{\permrank} \widetilde{\wtmatrix}^{(\ell)}$. Since every entry of every matrix $\widetilde{\wtmatrix}^{(\ell)}$ is non-negative, we have that every entry of $\widetilde{\wtmatrix}$ is also non-negative. We also claim that
\begin{align*}
\widetilde{\wtmatrix}_{ij} = \sum_{\ell=1}^{\permrank} \widetilde{\wtmatrix}^{(\ell)}_{ij} \stackrel{(i)}{\leq} \sum_{\ell=1}^{\permrank} {\wtmatrix}^{(\ell)}_{ij} = \wt_{ij} \leq 1,
\end{align*}
where the inequality (i) is a consequence of the set of inequalities~\eqref{EqnMonotoneApprox}. 
Thus we have that $\widetilde{\wtmatrix} \in [0,1]^{\numrows \times \numcols}$, $\wtmatrix \in [0,1]^{\numrows \times \numcols}$, and that the rank of $\widetilde{\wtmatrix}$ is at most $\permrank \widetilde{s}$. This result then yields the bound
\begin{align*}
\sum_{j=\permrank \widetilde{s} +1}^\numrows \sigma^2_j(\wt) \leq \frobnorm{ \wt - \widetilde{\wt} }^2 
\leq \sum_{i=1}^{\numrows} \sum_{j=1}^{\numcols} |\wt_{ij} - \widetilde{\wt}_{ij} | & =   \sum_{i=1}^{\numrows} \sum_{j=1}^{\numcols} | \sum_{\ell=1}^{\permrank} ( \wtmatrix^{(\ell)}_{ij} - \widetilde{\wtmatrix}^{(\ell)}_{ij} )| \\
&  \stackrel{(i)}{\leq} \sum_{i=1}^{\numrows} \sum_{j=1}^{\numcols} \sum_{\ell=1}^{\permrank} | \wtmatrix^{(\ell)}_{ij} - \widetilde{\wtmatrix}^{(\ell)}_{ij} |,
\end{align*}
where inequality $(i)$ follows from the triangle inequality. 
Now recall that every matrix $\widetilde{\wtmatrix}^{(\ell)}$ is obtained from Chatterjee's construction, and rewriting Chatterjee's result~\eqref{eq:STSVDChatterjeeRankOneAbs} for the matrices $\widetilde{\wtmatrix}^{(\ell)}$ presently under consideration, we obtain
\begin{align*}
\sum_{i=1}^\numrows | \widetilde{\wt}^{(\ell)}_{ij} - \wt^{(\ell)}_{ij}| \leq \frac{\numrows}{\widetilde{s}},
\end{align*}
for every $\ell \in [\permrank]$. 
As a consequence, we have
\begin{align*}
\sum_{j=\permrank \widetilde{s} +1}^\numrows \sigma^2_j(\wt) \leq \frac{\permrank \numrows \numcols}{\widetilde{s}} = \frac{\permrank^2 \numrows \numcols}{s},
\end{align*}
where we have substituted the relation $\widetilde{s} = \frac{s}{\permrank}$ to obtain the final result.

~\\ \noindent
Part (b): This result follows directly from the facts that the rank of $\wtmatrix$ is at most $\nnrank$, and the square of its Frobenius norm is at most $\numrows \numcols$.


\paragraph*{Proof of Lemma~\ref{LemWOp}}
Define an $((\numrows+\numcols)\times(\numrows+\numcols))$ matrix
$\noise''$ as
\begin{align*}
\noise'' = 
\frac{1}{\sqrt{\pp}}
\begin{bmatrix}
0 & \noise' \\ (\noise')^T & 0
\end{bmatrix}.
\end{align*}
From~\eqref{EqnDefnWprimePartialLowRankRepeat} and the construction
above, we have that the matrix $\noise''$ is symmetric, and has mutually
independent entries above the diagonal that have a mean of zero and a
variance upper bounded by $1$. Consequently, known results in random
matrix theory (e.g., see~\cite[Theorem 3.4]{chatterjee2014matrix}
or~\cite[Theorem 2.3.21]{tao2012topics}) yield the bound
$\opnorm{\noise''} \leq 2.01 \sqrt{\numrows + \numcols}$ with
probability at least $1 - e^{-\plaincon \max\{\numrows,
  \numcols\}}$, under the assumption $\pp \geq \frac{1}{\min\{\numrows,\numcols\}}\log^7(\numrows \numcols)$. One can also verify that
$\opnorm{\noise''} = \frac{1}{\sqrt{\pp}}\opnorm{\noise'}$, yielding
the claimed result.


\subsection{Proof of Proposition~\ref{PropAnyRank}}

We recall that for any integer $k \geq 0$, the notation
$\upperonesmx_{k}$ denotes an upper triangular matrix of size $(k
\times k)$ with all entries on and above the diagonal set as $1$, and
$\identitymx_{k}$ denotes the identity matrix of size $(k \times
k)$. Consider an $(\numrows \times \numcols)$
matrix $\wtmatrix$ with the following block structure:
\begin{align*}
\wtmatrix \defn \begin{bmatrix} \upperonesmx_{\nnrank - \permrank + 1}
  & 0 & 0\\ 0 & \identitymx_{\permrank - 1} & 0\\ 0 & 0 & 0
\end{bmatrix}.
\end{align*}
In the remainder of the proof, we show that $\fnnnrank{\wtmatrix} =
\nnrank$ and $\fnpermrank{\wtmatrix} = \permrank$. Using the ideas in
the construction of $\wtmatrix$ and the associated proof to follow,
one can construct many other matrices that have a non-negative rank of $\nnrank$ and a permutation-rank of $\permrank$, for any given value $1 \leq \permrank \leq \nnrank \leq
\min\{\numrows,\numcols\}$.

We partition the proof into four parts.\\

\noindent \underline{Proof of $\fnnnrank{\wtmatrix} \leq \nnrank$:}
One can write $\wtmatrix$ as a sum of $\nnrank$ matrices, each having
a non-negative rank of one: for each non-zero row, consider a
component matrix comprising that row and zeros
elsewhere. Consequently, we have $\fnnnrank{\wtmatrix} \leq
\nnrank$. \\

\noindent \underline{Proof of $\fnnnrank{\wtmatrix} \geq \nnrank$:}
Observe that the
(conventional) rank of $\wtstar$ equals $\nnrank$. Since the rank of any
matrix is a lower bound on its non-negative rank, we have that
$\fnnnrank{\wtmatrix} \geq \nnrank$. We have thus established that the non-negative rank of this
matrix equals exactly $\nnrank$. \\

\noindent \underline{Proof of $\fnpermrank{\wtmatrix} \leq
  \permrank$:} Observe that the $(\numrows \times \numcols)$ matrix
with $\upperonesmx_{\nnrank - \permrank + 1}$ as its top-left
submatrix and zeros elsewhere has a permutation-rank of $1$. Moreover,
any $(\numrows \times \numcols)$ matrix with exactly one entry as $1$
and the remaining entries $0$ also has a permutation-rank of $1$, and
hence a $(\numrows \times \numcols)$ matrix with
$\identitymx_{\permrank - 1}$ as its submatrix and zeros elsewhere has
a permutation-rank of at most $(\permrank-1)$. Putting these arguments
together, we obtain the bound $\fnpermrank{\wtmatrix} \leq
\permrank$.\\

\noindent \underline{Proof that $\fnpermrank{\wtmatrix} \geq
  \permrank$:} 
%
First observe that the matrix
\begin{align*}
\identitymx_{2 \times 2} =
\begin{bmatrix}
1 & 0 \\ 0 & 1
\end{bmatrix}
\end{align*}
does not belong to $\permset(1)$. It follows that any matrix containing $\identitymx_{2 \times 2}$ as a
submatrix cannot belong to the set $\permset(1)$. It further follows
that for any positive integer $k$, the matrix $\identitymx_{k \times
  k}$ must have a permutation rank of at least $k$. Finally, observe
that the matrix $\wtmatrix$ contains
$\identitymx_{\permrank \times \permrank}$ as its submatrix (given by
the intersection of rows $\{\nnrank - \permrank, \ldots, \nnrank\}$
with the columns $\{\nnrank - \permrank, \ldots, \nnrank\}$). It
follows that $\wtmatrix$ must have a permutation rank of at least
$\permrank$, thereby proving the claim.


\subsection{Proof of Proposition~\ref{PropInclusion}}

We assume for ease of exposition that $\numrows$ and $\numcols$ are
divisible by $\temprank$. Otherwise, since $\temprank \leq \half
\min\{\numrows, \numcols\}$, one may take floors or ceilings which
will change the result only by a constant factors. Since $\nnset(\temprank) \subseteq \permset(\temprank)$, we have $\sup \limits_{\wt
	\in \nnset(\temprank) } \ \inf \limits_{\wt' \in \permset(\temprank) } \frobnorm{\wt - \wt'}^2 = 0$. In what follows, we show that $\sup \limits_{\wt
	\in \permset(\temprank)} \ \inf \limits_{\wt' \in \nnset(\temprank)} \frobnorm{\wt - \wt'}^2 \geq \frac{\plaincon \numrows \numcols }{ \temprank}$.

Consider the block matrix $\wttil \in
[0,1]^{\frac{\numrows}{\temprank} \times \frac{\numcols}{\temprank}}$:
\begin{align}
\label{EqnRankOneBadApprox}
\wttil = 
\begin{bmatrix}
1 & 1 \\ 1 & 0
\end{bmatrix},
\end{align}
where each of the four blocks is of size $(\frac{\numrows}{2 \temprank}
\times \frac{\numcols}{2 \temprank})$. The following lemma shows that
the best rank-$1$ approximation to $\wttil$ has a large
approximation error:
\begin{lemma}
\label{LemRankOneBadApprox}
For the matrix $\wttil$ defined in~\eqref{EqnRankOneBadApprox}, for
any vectors $\lowLeftVec{} \in \reals^{\numrows}$ and $\lowRightVec{}
\in \reals^{\numcols}$, it must be that
\begin{align*}
\frobnorm{ \wttil - \lowLeftVec{} {\lowRightVec{}}^T }^2 \geq
\plaincon \frac{\numrows \numcols}{\temprank^2},
	\end{align*}
where $\plaincon > 0$ is a universal constant.
\end{lemma}

We now use the matrix $\wttil$ defined in~\eqref{EqnRankOneBadApprox}
to build the following block matrix $\wtmatrix \in
\permset(\temprank)$:
\begin{align*}
\wtmatrix \defn \begin{bmatrix} \wttil & 0 & \cdots & 0 \\ 0 &
  \wttil & \cdots & 0 \\ \vdots & \vdots & \ddots & \vdots \\ 0 & 0
  & \cdots & \wttil
\end{bmatrix}.
\end{align*}
In words, the matrix $\wtmatrix$ is a block-diagonal matrix where the
diagonal has $\temprank$ copies of $\wttil$.

Due to the block diagonal structure of $\wtmatrix$, the singular
values of $\wtmatrix$ are simply $\temprank$ copies of the singular
values of its constituent matrix $\wttil$. Consequently, we have
that for any matrix $\wtmatrix' \in \nnset(\temprank)$:
\begin{align*}
\frobnorm{ \wtmatrix - \wtmatrix'}^2 \geq \temprank (\frobnorm{\wttil}^2
- \opnorm{\wttil}^2) \stackrel{(i)}{\geq} \plaincon \frac{\numrows
  \numcols}{\temprank},
\end{align*} 
as claimed, where the inequality $(i)$ is a consequence of
Lemma~\ref{LemRankOneBadApprox}.


\paragraph*{Proof of Lemma~\ref{LemRankOneBadApprox}}

Consider any value $i \in [\frac{\numrows}{2\temprank}]$ and $j \in
[\frac{\numcols}{2\temprank}]$.  Then we claim that
\begin{align}
(\wttil_{i,j} - [\lowLeftVec{} {\lowRightVec{}}^T]_{i,j})^2 +
  (\wttil_{i + \frac{\numrows}{2\temprank}, j} - [\lowLeftVec{}
    {\lowRightVec{}}^T]_{i + \frac{\numrows}{2\temprank}, j })^2 + &
  (\wttil_{i, j + \frac{\numcols}{2\temprank}} - [\lowLeftVec{}
    {\lowRightVec{}}^T]_{i, j + \frac{\numcols}{2\temprank}})^2
  \nonumber\\
& + (\wttil_{i + \frac{\numrows}{2\temprank}, j +
    \frac{\numcols}{2\temprank}} - [\lowLeftVec{} {\lowRightVec{}}^T]_{i
    + \frac{\numrows}{2\temprank}, j + \frac{\numcols}{2\temprank}})^2
  \geq 0.01.
\label{EqnRankOneBadApproxProof}
\end{align}
If not, then for the choice of $\wttil$
in~\eqref{EqnRankOneBadApprox}, we must have $[\lowLeftVec{}
  {\lowRightVec{}}^T]_{i,j} \in (0.9,1.1)$, $[\lowLeftVec{}
  {\lowRightVec{}}^T]_{i + \frac{\numrows}{2\temprank}, j } \in
(0.9,1.1)$, $[\lowLeftVec{} {\lowRightVec{}}^T]_{i, j +
  \frac{\numcols}{2\temprank}} \in (0.9,1.1)$ and $[\lowLeftVec{}
  {\lowRightVec{}}^T]_{i + \frac{\numrows}{2\temprank}, j +
  \frac{\numcols}{2\temprank}} < 0.1$. However, since $[\lowLeftVec{}
  {\lowRightVec{}}^T]_{i',j'} = \lowLeftVec{}_{i'}
\lowRightVec{}_{j'}$ for every coordinate $(i',j')$, we also have
\begin{align*}
[\lowLeftVec{} {\lowRightVec{}}^T]_{i,j} \times [\lowLeftVec{}
  {\lowRightVec{}}^T]_{i + \frac{\numrows}{2\temprank}, j +
  \frac{\numcols}{2\temprank}} = [\lowLeftVec{} {\lowRightVec{}}^T]_{i +
  \frac{\numrows}{2\temprank}, j } \times [\lowLeftVec{}
  {\lowRightVec{}}^T]_{i, j + \frac{\numcols}{2\temprank}},
\end{align*}
which contradicts the required ranges of the individual coordinates.
Summing the bound~\eqref{EqnRankOneBadApproxProof} over all values of
$i \in [\frac{\numrows}{2\temprank}]$ and $j \in
[\frac{\numcols}{2\temprank}]$ yields the claimed result.


\subsection{Proof of Proposition~\ref{PropNoConvex}}

Consider any set $\genclass$ and any convex set $\convexclass$.  We
begin with a key lemma that establishes a relation between $\hausDis(\genclass, \convexclass)$ and a proposed
notion of the inherent convexity of $\genclass$.
\begin{lemma}
For any set $\genclass \subseteq [0,1]^{\numrows \times \numcols}$ and
any convex set $\convexclass\subseteq [0,1]^{\numrows \times   \numcols}$, it must be that
\begin{align}
\hausDis(\genclass, \convexclass) & \geq \frac{2}{9} \sup \limits_{\wt_1
  \in \genclass,\ \wt_2 \in \genclass} ~ \inf \limits_{\wt_0 \in
  \genclass} \frobnorm{\frac{1}{2} (\wt_1 + \wt_2) - \wt_0}^2.
	\label{EqnGenConvex}
\end{align}	
	\label{LemNoconvex}
\end{lemma}
The left hand side of inequality~\eqref{EqnGenConvex} is the Hausdorff
distance between the sets $\genclass$ and $\convexclass$ in terms of
the squared Frobenius norm. The right hand side of the inequality
represents a notion of the inherent convexity of the set $\genclass$.

With this lemma in place, we now complete the remainder of the
proof. To this end, we set $\genclass = \permone$, and let
$\convexclass$ be any convex set of $[0,1]$-valued $(\numrows \times
\numcols)$ matrices.

We now construct a pair of matrices $\wt_1 \in \permone$ and $\wt_2
\in \permone$ that we use to lower bound the right hand side
of~\eqref{EqnGenConvex}. Define matrices $\wt_1 \in \permone$ and $\wt_2 \in \permone$ as
\begin{align*}
[\wt_1]_{ij} = 
\begin{cases}
1 & \mbox{if } i \leq \frac{\numrows}{2},\ j \leq
\frac{\numcols}{2}\\ 0 & \mbox{otherwise},
\end{cases} \qquad \text{and} \qquad
[\wt_2]_{ij} =
\begin{cases}
1 & \mbox{if } i > \frac{\numrows}{2},\ j > \frac{\numcols}{2}\\ 0 &
\mbox{otherwise}.
\end{cases}
\end{align*}
It follows that the entries of the matrix $\frac{1}{2}(\wt_1 + \wt_2)$
are given by:
\begin{align*}
[\frac{1}{2}(\wt_1 + \wt_2)]_{ij} = 
\begin{cases}
\frac{1}{2} & \mbox{if } (i \leq \frac{\numrows}{2},\ j \leq
\frac{\numcols}{2}) \mbox{~or~} (i > \frac{\numrows}{2},\ j >
\frac{\numcols}{2}) \\ 0 & \mbox{otherwise}.
\end{cases}
\end{align*}

Now consider any pair of integers $(i,j) \in [\lfloor \numrows/2
  \rfloor] \times [\lfloor \numcols/2 \rfloor ]$. Then the $(2 \times
2)$ submatrix of $\frac{1}{2}(\wt_1 + \wt_2)$ formed by its entries
$(i,j)$, $(i + \lceil \numrows/2 \rceil, j)$, $(i, j+ \lceil
\numcols/2 \rceil)$ and $(i + \lceil \numrows/2 \rceil, j+ \lceil
\numcols/2 \rceil)$ equals
\begin{align*}
\begin{bmatrix}
\half & 0 \\ 0 & \half
\end{bmatrix}.
\end{align*}
It is easy to verify that there is a constant $\plaincon>0$ such that
the squared Frobenius norm distance between this rescaled identity
matrix and any $(2 \times 2)$ matrix in $\permone$ is at least
$\plaincon$. Since this argument holds for any choice of $(i,j) \in
[\lfloor \numrows/2 \rfloor] \times [\lfloor \numcols/2 \rfloor ]$,
summing up the errors across each of these sets of entries yields
\begin{align*}
\frobnorm{\half (\wt_1 + \wt_2) - \wt}^2 \geq \plaincon' \numrows
\numcols, \qquad \mbox{for every matrix $\wt \in \permone$},
\end{align*}
where $\plaincon' > 0$ is a universal constant. Finally, substituting
this bound in Lemma~\ref{LemNoconvex} yields the claimed result.

It remains to prove Lemma~\ref{LemNoconvex}.


\paragraph{Proof of Lemma~\ref{LemNoconvex}.}
%

Consider any matrices $\wt_1 \in \genclass$ and $\wt_2 \in
\genclass$. From the definition of the Hausdorff distance $\hausDis$, we know that there
exist matrices $\wttil_1 \in \convexclass$ and $\wttil_2 \in
\convexclass$ such that
\begin{align}
\label{EqnHaus1}
\frobnorm{\wt_i - \wttil_i}^2 \leq \hausDis(\genclass, \convexclass),
\quad \mbox{for } i \in \{1,2\}.
\end{align}
Since $\convexclass$ is a convex set, we also have $\half (\wttil_1 +
\wttil_2)\in \convexclass$. Then from the definition of $\hausDis$, we
also know that there exists a matrix $\wt_0 \in \genclass$ such that
\begin{align}
\label{EqnHaus2}
\frobnorm{\half (\wttil_1 + \wttil_2) - \wt_0}^2 \leq
\hausDis(\genclass, \convexclass).
\end{align} 
Finally, applying the triangle inequality to the bounds~\eqref{EqnHaus1} and~\eqref{EqnHaus2} yields
\begin{align*}
\frobnorm{\half(\wt_1 + \wt_2) - \wt_0 }^2 &\leq
3\frobnorm{\half(\wttil_1 + \wttil_2) - \wt_0 }^2 + \frac{3}{4}
\frobnorm{\wt_1 - \wttil_1}^2 + \frac{3}{4} \frobnorm{\wt_2 -
  \wttil_2}^2 \\
& \leq \frac{9}{2} \hausDis(\genclass, \convexclass).
\end{align*}


\subsection{Proof of Proposition~\ref{PropUniqueDecomposition}}

Suppose there exists a coordinate pair $(i,j)$ such the stated
condition is violated. Then there must exist two distinct values
$\ell_1 \in [\fnpermrank{\wt}]$ and $\ell_2 \in [\fnpermrank{\wt}]$ that satisfy the
following three conditions:
\begin{enumerate}[label=(\alph*),topsep=0pt,itemsep=-1ex,partopsep=1ex,parsep=1ex]
\item $\wt^{(\ell_1)}_{ij} > 0$ and
$\wt^{(\ell_2)}_{ij} > 0$,
\item The value of $\wt^{(\ell_1)}_{ij}$ is different
from all other entries in $\wt^{(\ell_1)}$, and
\item
The value of $\wt^{(\ell_2)}_{ij}$ is different from all other entries in
$\wt^{(\ell_2)}$.
\end{enumerate}
In addition, the fact that $\wt^{(\ell_1)}_{ij} + \wt^{(\ell_2)}_{ij}
\in (0,1)$ for every coordinate $(i,j)$, along with condition (a)
above, imply a fourth condition:
\begin{enumerate}[label=(\alph*),topsep=0pt,itemsep=-1ex,partopsep=1ex,parsep=1ex]
	\item[(d)] $\wt^{(\ell_1)}_{ij} < 1$
and $\wt^{(\ell_2)}_{ij} < 1$.
\end{enumerate}

Now for any $\epsilon > 0$, define $\wttil^{(\ell_1)}_\epsilon$ and $\wttil^{(\ell_2)}_\epsilon$ to be matrices obtained by replacing  the
$(i,j)^{th}$ entries of the matrices $\wt^{(\ell_1)}$ and
$\wt^{(\ell_2)}$ with $(\wt^{(\ell_1)}_{ij} + \epsilon)$ and
$(\wt^{(\ell_2)}_{ij} - \epsilon)$ respectively.
Now, conditions (b)--(d) in tandem imply that there exists some value
$\epsilon > 0$ such that \emph{all} of the following properties
hold:
\begin{enumerate}[label=(\roman*),topsep=0pt,itemsep=-1ex,partopsep=1ex,parsep=1ex]
\item $[\wt^{(\ell_1)}_\epsilon]_{ij} \in [0,1]$,
\item $[\wt^{(\ell_2)}_\epsilon]_{ij}  \in [0,1]$, and 
\item The relative ordering of the entries of $\wt^{(\ell_1)}$ is identical to the relative ordering of the entries of $\wt^{(\ell_1)}_\epsilon$; the relative ordering of the entries of $\wt^{(\ell_1)}$ is identical to the relative ordering of the entries of $\wt^{(\ell_1)}_\epsilon$.
\end{enumerate}
Properties (i) and (ii) imply that $\wttil^{(\ell_1)}_\epsilon \in [0,1]^{\numrows \times \numcols}$ and $\wttil^{(\ell_2)}_\epsilon \in [0,1]^{\numrows \times \numcols}$. Combined with property (iii), we also have $\wttil^{(\ell_1)}_\epsilon \in \permone$ and $\wttil^{(\ell_2)}_\epsilon \in \permone$. Finally, from the construction of the matrices $\wttil^{(\ell_1)}_\epsilon$ and $\wttil^{(\ell_2)}_\epsilon$, it is easy to see the relation 
\begin{align*}
\wt = \wttil^{(\ell_1)}_\epsilon + \wttil^{(\ell_2)}_\epsilon + \sum_{i \in [ \fnpermrank{\wt}] \backslash \{\ell_1,\ell_2\}} \wt^{(i)}.
\end{align*} 
This decomposition of $\wt$ is a different, valid permutation-rank decomposition of $\wt$. 


\section{Discussion and future work}
\label{SecConclusion}

We posit that the conventional low-rank models for matrix completion
and denoising are equivalent to ``parametric'' assumptions with
undesirable implications. We propose a new permutation-rank approach
and argue, by means of a philosophical discussion as well as
theoretical guarantees, that this approach offers significant benefits
at little additional cost. Our work also contributes to a growing body of literature~\cite[Part~1]{shah2017thesis},~\cite{shah2015stochastically, shah2015simple, shah2016permutation, shah2016feeling,heckel2016active, chatterjee2016estimation, flammarion2016optimal, chen2017competitive} on moving towards more flexible models based on permutations
that provide robustness to model mismatches.

Our work gives rise to some useful open problems that we hope to
address in future work. In this paper, we established benefits of the
permutation-based approach for the matrix completion problem under the
random design observation setting. In the literature, the classical
low (non-negative) rank matrix completion problem is more recently also
studied under other observation models such as weighted random
sampling~\cite{negahban2012restricted}, fixed
design~\cite{jain2013low, klopp2014noisy}, streaming/active
learning~\cite{yun2015streaming, jin2016provable, balcan2016noise}, or
biased observation models~\cite{hsieh2015pu}, which are also of
interest in the context of permutation-rank matrix completion. A
second open problem is to close the gap between the statistically
optimal minimax rate of estimation 
and the best known rate 
for polynomial-time computable algorithms for the permutation-rank
model. Any solution to this problem may also contribute to the
understanding of some other open problems in the literature (e.g.,
see~\cite{shah2015stochastically,flammarion2016optimal,shah2016permutation})
on the gap between the statistical and computational aspects of
estimation under an unknown permutation.


\subsection*{Acknowledgments}

The work of MJW and NBS was partially supported by DOD Advanced
Research Projects Agency W911NF-16-1-0552 and National Science
Foundation grant NSF-DMS-1612948. The work of SB was supported by 
NSF grant DMS-1713003.


~\\
\appendix

\noindent{\Large \bf Appendix}

\section{Intuitive algorithms that provably fail}
\label{SecAlgoFail}

In this section, we present two intuitive polynomial-time computable
algorithms for the permutation-rank setting---one for estimating
$\wtstar$ from $\obs$, and one for decomposing $\wtstar$ into its
constituent permutation-rank-one matrices---and show that these
algorithms provably fail.  Our goal in describing these negative
results is as a complement to the positive results provided in the
main text, and with the hope that the points of failure of these
algorithms may form starting points for subsequent research.


\subsection{An intuitive polynomial-time estimator for $\wtstar$ from $\obs$}

In this section, we consider the problem of estimating the matrix
$\wtstar$ from noisy and partial observations $\obs$ as defined
earlier in equation~\eqref{EqnDefnObsLowrank}. For simplicity, we
assume that $\pp=1$.  For any vector $z \in \reals^m$, we let vector $z_+ \in \reals^m$ with entries $[z_+]_i = \max\{z_i, 0\}$ represent the positive component of $z$, and vector $z_- \in \reals^m$ with entries $[z_-]_i = \min\{z_i, 0\}$ represent the negative component of $z$. 
 
We first provide some intuition and background to motivate the estimator we study in this section, and then present a formal definition. Denote the permutation-rank of matrix $\wtstar$ as $\permstar \defn \fnpermrank{\wtstar}$, and assume that the value of $\permstar$ is known. The goal is to obtain an estimate $\wthat \in \permset(\permstar)$ from the observed matrix $\obs$ such that the error $\frac{1}{\numrows \numcols} \frobnorm{\wtstar - \wthat}^2$ is as small as possible. For any such matrix $\wthat$, let us use the following notation for its permutation-rank decomposition: $\wthat = \sum_{\ell = 1}^{\permstar} \wthat^{(\ell)}$ where $\wthat^{(\ell)} \in \permone$ for every $\ell \in [\permstar]$. Further, we let $\widehat{\pi}^{(\ell)}$ and
$\widehat{\sigma}^{(\ell)}$ respectively denote the permutation of the rows and columns of $\wthat^{(\ell)}$. 


Past literature on computationally-efficient estimation for such problems provides us with estimators with the following two distinct goals: (i)  to estimate the permutations of the constituent matrices in the permutation-rank decomposition, and (ii) estimators to compute the entries of the constituent matrices given the permutations. For each of these two goals, we describe a natural estimator below from past works.\\

\emph{Estimating the permutations via singular value decomposition:} Compute the singular value
decomposition $\obs = \sum_{\ell=1}^{\min\{\numrows,\numcols\}}
\leftvec^{(\ell)} [\rightvec^{(\ell)}]^T$  such that the vectors $\{\leftvec^{(1)},\ldots,\leftvec^{(\min\{\numrows,\numcols\})}\}$ are mutually orthogonal, the vectors $\{\rightvec^{(1)},\ldots,\rightvec^{(\min\{\numrows,\numcols\})}\}$ are mutually orthogonal, and $[\leftvec^{(1)}]^T \rightvec^{(1)} \geq \ldots \geq [\leftvec^{(\min\{\numrows,\numcols\})}]^T \rightvec^{(\min\{\numrows,\numcols\})}$. In order to resolve a global sign ambiguity, we also mandate the condition $\Lnorm{\leftvec^{(\ell)}_+}{2} \geq \Lnorm{\leftvec^{(\ell)}_-}{2}$ for every $\ell \in [\min\{\numrows,\numcols\}]$.  Finally, for each $\ell \in
[\permstar]$, set $\widehat{\pi}^{(\ell)}$ and
$\widehat{\sigma}^{(\ell)}$ as the ordering of
the entries of $\leftvec^{(\ell)}$ and $\rightvec^{(\ell)}$ respectively.

From past works~\cite{shah2016permutation}, this estimator for the permutations is known to possess appealing properties for the case when $\permstar= 1$. For instance, it is not hard to see that in a noiseless setting where $\obs = \wtstar$, the estimator will yield exactly the row and column permutations of $\wtstar \in \permone$. This fact
is employed in the paper~\cite{shah2016permutation} to obtain
consistent estimates of the permutations associated to an unknown matrix in $\permone$ in the context of a ``crowd labeling'' problem. It is also not hard to verify that the estimator is can be computed in a computationally-efficient manner. 
\\

\emph{Estimating the entries via least squares, when given the permutations:} Given some estimate $\widehat{\pi}^{(1)},\widehat{\sig}^{(1)},\ldots,\widehat{\pi}^{(\permstar)},\widehat{\sig}^{(\permstar)}$ of the permutations associated to the permutation-rank decomposition of $\wtstar$, the following estimator $\wthat$ provides an estimate of the matrix $\wtstar$ as well as the matrices in its permutation-rank decomposition. 
\begin{align}
\wthat \in & \argmin_{\wtmatrix \in [0,1]^{\numrows \times \numcols} } \frobnorm{ \obs -
  \wtmatrix}^2 \label{EqnLSEDoesntWork} \\ & \mbox{such that}\quad
\wtmatrix = \sum_{\ell=1}^{\permstar} \widetilde{\wtmatrix}^{(\ell)}, \text{ and}
\nonumber \\
& \phantom{such that}\quad \mbox{$\widetilde{\wtmatrix}^{(\ell)} \in
  \permone$ with rows and columns ordered by $(\widehat{\pi}^{(\ell)},
  \widehat{\sigma}^{(\ell)})$, for every $\ell \in
          [\permstar]$}. \nonumber
\end{align} 

The aforementioned estimator $\wthat$ is a natural extension of the least-squares estimators studied in past works~\cite{shah2015stochastically, shah2016feeling} for the case when $\wtstar \in \permone$. The estimator is known to have appealing properties from both the statistical and computational perspectives. From a computational standpoint, all of the constraints in the optimization
program~\eqref{EqnLSEDoesntWork} can be expressed as a set of
(polynomial number of) linear inequalities, thereby making the
optimization problem computationally tractable. From a statistical standpoint, if the given permutations $(\widehat{\pi}^{(1)},\widehat{\sig}^{(1)},\ldots,\widehat{\pi}^{(\permstar)},\widehat{\sig}^{(\permstar)})$ are exactly (or approximately) equal to the permutations associated to a permutation-rank decomposition of $\wtstar$, the proofs of the results in~\cite{shah2015stochastically,shah2016feeling} as well as Theorem~\ref{ThmStatOracle} in the present paper imply that the estimator $\wthat$ is minimax optimal for estimating $\wtstar$ from $\obs$. The estimator $\wthat$ continues  to remain statistically efficient if the permutations are known up to a reasonable approximation.\\

Given the two intuitive estimators discussed above, a natural means to estimate $\wtstar$ from $\obs$ is to concatenate these two estimators to obtain the following two-step estimator:\\
Step 1: From $\obs$, obtain an estimate $(\widehat{\pi}^{(1)},\widehat{\sig}^{(1)},\ldots,\widehat{\pi}^{(\permstar)},\widehat{\sig}^{(\permstar)})$ of the permutations of the decomposition of $\wtstar$ via the singular value decomposition-based estimator described above.\\
Step 2: Using the estimates of the permutations, obtain an estimate $\wthat$ of $\wtstar$ via the least squares projection~\eqref{EqnLSEDoesntWork}.\\

We
believe that when $\permstar=1$, this estimator is not only
consistent, but it has an expected error decaying at the rate
$\Order(\min\{\numrows,\numcols\}^{-1/2})$. We now show that in fact as soon as one moves
to the setting of $\permstar > 1$, this estimator is no longer even
consistent---even if there is no noise.
\begin{proposition}
\label{PropBreakLS}
There exists a matrix $\wtstar \in
\permset(2)$ such that when $\obs = \wtstar$, the two-step estimator
$\wthat$ has an error lower
bounded as
\begin{align*}
\frac{1}{\numrows \numcols} \frobnorm{\wtstar - \wthat}^2 \geq \ULOW,
\end{align*}
with probability $1$.
\end{proposition}

The proof of this result is provided in Appendix~\ref{AppProofPropBreakLS}. The proof also demonstrates an identical negative result
for the following modified estimation algorithm: In computing $\wthat$ as above, instead of taking
only the permutations of the top $\permstar$ singular vectors, collect
permutations from singular vectors until you obtain $\permstar$
\emph{distinct} permutations; then apply the least squares projection step to
these $\permstar$ distinct permutations.


\subsection{An intuitive greedy algorithm for permutation-rank decomposition}

Consider any matrix any matrix $\wt \in [0,1]^{\numrows \times
  \numcols}$. The singular value decomposition of $\wt$ into components having a (conventional) rank of one can be performed with the following greedy
algorithm:
\begin{itemize}
	\itemsep0em
\item Let $\widehat{\temprank} = 1$
\item While $\wt \neq \sum_{\ell=1}^{\widehat{\temprank} - 1}
          \wthat^{(\ell)}$:
\begin{itemize}
\item Let $\wthat^{(\widehat{\temprank})} \in \argmin
  \limits_{\substack{\wt' = \leftvec \rightvec^T\\(\leftvec,\rightvec) \in
      \reals^\numrows \times \reals^\numcols}}
  \matsnorm{ \wt - \sum_{\ell=1}^{\widehat{\temprank}
      - 1} \wthat^{(\ell)} - \wt'}{F}$
\item $\widehat{\temprank} = \widehat{\temprank} + 1$
\end{itemize}
\item Output $\widehat{\temprank}$ as the rank of $\wt$ and
  $\{\wthat^{(1)}, \ldots, \wthat^{(\widehat{\temprank})} \}$ as its
  singular value decomposition.
\end{itemize}
An obvious question that arises is whether a similar greedy algorithm
works to obtain a permutation-rank decomposition.

To this end, consider any value $\entrynorm \geq 1$, and for any
matrix $\wt$, let $\matsnorm{\wt}{\entrynorm}$ denote its entry-wise
norm $\matsnorm{\wt}{\entrynorm} \defn \big(\sum_{i,j}
(\wt_{ij})^\entrynorm \big)^{\frac{1}{\entrynorm}}$. Then the natural
analogue of the aforedescribed algorithm in the context of
permutation-rank decomposition is as follows:
\begin{itemize}
\item Let $\widehat{\permrank} = 1$
\item While $\wt \neq \sum_{\ell=1}^{\widehat{\permrank} - 1} \wthat^{(\ell)}$:
  \begin{itemize}
  \item Let $\wthat^{(\widehat{\permrank})} \in \argmin \limits_{\wt' \in \permset(1)} \matsnorm{ \wt - \sum_{\ell=1}^{\widehat{\permrank} - 1} \wthat^{(\ell)} - \wt'}{\entrynorm}$
  \item $\widehat{\permrank} = \widehat{\permrank} + 1$
  \end{itemize}
\item Output $\widehat{\permrank}$ as the permutation-rank of $\wt$
  and $\{\wthat^{(1)}, \ldots, \wthat^{(\widehat{\permrank})} \}$ as its
  permutation-decomposition
\end{itemize}

The following proposition investigates whether such an algorithm will work.
\begin{proposition}\label{PropBreakGreedyDecomp}
For any values of $\numrows$, $\numcols$ and $\permrank \geq 2$,
there exists an $(\numrows \times \numcols)$ matrix $\wt \in
\permset(\permrank)$ such that the above algorithm outputs a
decomposition of permutation-rank at least $(\permrank + 1)$.
\end{proposition}

The guaranteed incorrectness of the permutation rank of the output of
the algorithm also directly implies that the decomposition is also
incorrect.


\subsection{Proofs}

We now present the proofs of the negative results introduced in this
section.


\subsubsection{Proof of Proposition~\ref{PropBreakLS}}
\label{AppProofPropBreakLS}

In what follows, for clarity of exposition, we ignore issues
pertaining to floors and ceilings of numbers, as they affect the
results only by a constant factor.

We begin by defining a matrix $\wtstar \in \permset(2)$ as
\begin{align*}
\wtstar = \wtmatrix^{(1)} + \wtmatrix^{(2)},
\end{align*}
with
\begin{align*}
\wtmatrix^{(1)} = \leftvec^{(1)} (\rightvec^{(1)})^T + \leftvec^{(2)} (\rightvec^{(2)})^T
\mbox{\quad and \quad} \wtmatrix^{(2)} = \leftvec^{(3)} (\rightvec^{(3)})^T.
\end{align*}
Set
\begin{align*}
\begin{matrix}
\leftvec^{(1)} & = & [1 & .9 ~ \cdots ~ .9 ~ & .8 ~ \cdots ~ .8 ~&0 ~ \cdots
  ~ 0 &]^T \\ \leftvec^{(2)} &=& [0 & .2 ~ \cdots ~ .2 ~ & -.1 ~ \cdots ~
  -.1~ &0 ~ \cdots ~ 0 &]^T\\ \leftvec^{(3)} & =& [0 & \underbrace{0 ~ \cdots
    ~ 0}_{\subveclength_1 (\numrows-1)} ~& \underbrace{0 ~ \cdots ~
    0}_{\subveclength_2 (\numrows-1)} ~ & \underbrace{1 ~ \cdots ~
    1}_{\subveclength_3 (\numrows-1)}&]^T,
\end{matrix}
\end{align*}
and
\begin{align*}
\begin{matrix}
\rightvec^{(1)} &=& [1 & .9 ~ \cdots ~ .9 ~ & .8 ~ \cdots ~ .8 ~&0 ~ \cdots ~
  0 &]^T \\ \rightvec^{(2)} &=& [0 & .2 ~ \cdots ~ .2 ~ & -.1 ~ \cdots ~ -.1 ~
  &0 ~ \cdots ~ 0 &]^T\\ \rightvec^{(3)} & =& [0 & \underbrace{0 ~ \cdots ~
    0}_{\subveclength_1 (\numcols-1)} ~& \underbrace{0 ~ \cdots ~
    0}_{\subveclength_2 (\numcols-1)} ~ & \underbrace{1 ~ \cdots ~
    1}_{\subveclength_3 (\numcols-1)}&]^T,
\end{matrix}
\end{align*}
where $\subveclength_1 = .684$, $\subveclength_2 = .304$, and
$\subveclength_3 = .012$. It is easy to verify that all entries of the
matrices $\wtstar, \wtmatrix^{(1)}, \wtmatrix^{(2)}$ lie in the
interval $[0,1]$ and that $\wtmatrix^{(1)} \in \permone$ and
$\wtmatrix^{(2)} \in \permone$ and $\wtstar \in \permset(2) \backslash
\permset(1)$.

One can further verify the following properties of this construction:
\begin{enumerate}
\item $\inprod{\leftvec^{(\ell)} }{\leftvec^{(\ell')}} = 0$ and $\inprod{\rightvec^{(\ell)}
}{\rightvec^{(\ell')}} = 0$ for every $\ell \neq \ell' \in
  \{1,2,3\}$.\label{PropNegativeConstructionOrthogonal}
\item $\Lnorm{\leftvec^{(1)}}{2} > \Lnorm{\leftvec^{(2)}}{2}> \Lnorm{\leftvec^{(3)}}{2}$
  and $\Lnorm{\rightvec^{(1)}}{2} > \Lnorm{\rightvec^{(2)}}{2} >
  \Lnorm{\rightvec^{(3)}}{2}$.\label{PropNegativeConstructionMaxVec}
\item The (conventional) rank of $\wtstar$ is
  $3$.\label{PropNegativeConstructionRanks}
\item $\leftvec^{(1)}$, $\leftvec^{(2)}$ and $\leftvec^{(3)}$ have different permutations
  of their entries; likewise, $\rightvec^{(1)}$, $\rightvec^{(2)}$ and $\rightvec^{(3)}$ have
  different permutations of their
  entries.\label{PropNegativeConstructionDiffPerm}
  \item $\Lnorm{\leftvec^{(\ell)}_+}{2} \geq \Lnorm{\leftvec^{(\ell)}_-}{2}$ and $\Lnorm{\rightvec^{(\ell)}_+}{2} \geq \Lnorm{\rightvec^{(\ell)}_-}{2}$ for every $\ell \in [3]$. \label{ProfNegativeConstructionSign}
\end{enumerate}

The five properties listed above imply that the following decomposition of $\wtstar$,
\begin{align*}
\wtstar = \sum_{\ell=1}^{3} \leftvec^{(\ell)} (\rightvec^{(\ell)})^T,
\end{align*}
is a valid singular value decomposition with the global signs of the constituent vectors satisfying the conditions of Step 1 of the
algorithm. Consequently, the $\permstar = 2$ estimated permutations in
Step 1 of the algorithm are those given by the respective orderings of the entries of the vectors $\{\leftvec^{(1)},\rightvec^{(1)}\}$ and
$\{\leftvec^{(2)},\rightvec^{(2)}\}$.

Observe that the entries $2$ to $(1+\subveclength_1
(\numrows-1))$ of both $\leftvec^{(1)}$ and $\leftvec^{(2)}$ have values higher than
the last $\subveclength_3 (\numrows-1)$ entries of these vectors, and hence
this ordering is reflected in the respective permutations
derived from these vectors. Likewise, the entries $2$ through
$(1+\subveclength_1 (\numcols-1))$ are ranked higher than the last
$\subveclength_3 (\numcols-1)$ entries in the permutation derived from
the vectors $\rightvec^{(1)}$ and $\rightvec^{(2)}$. Due to this collection of
inequalities, the least squares program in Step 2 of the algorithm
must mandate that
\begin{align}
\label{EqnNegativeConstructionInequalities}
\wtmatrix_{ij} \geq \wtmatrix_{i'j'} \geq \wtmatrix_{i''j''},
\end{align}
whenever $2 \leq i,i' \leq 1 + \subveclength_1(\numrows-1)$; $\numrows
- \subveclength_3(\numrows-1) < i'' \leq \numrows$; $2 \leq j \leq
1+\subveclength_1 (\numcols - 1)$; and $ \numcols -
\subveclength_3(\numcols-1) < j', j'' \leq \numcols$.  However, for
each coordinate in this range, we also have $\wtstar_{ij} = .85,
\wtstar_{i'j'} = 0, \wtstar_{i''j''} = 1$. Consequently, any triplet
of values $(\wtmatrix_{ij}, \wtmatrix_{i'j'}, \wtmatrix_{i''j''})$
that follows the ordering~\eqref{EqnNegativeConstructionInequalities}
must necessarily incur an error lower bounded as
\begin{align*}
(\wt_{ij} - \wtstar_{ij})^2 + (\wt_{i'j'} - \wtstar_{i'j'})^2 +
  (\wt_{i''j''} - \wtstar_{i''j''})^2 \geq \plaincon,
\end{align*}
for some universal constant $\plaincon > 0$. Summing up the errors
over all the entries of the matrix in the aforementioned coordinate
set yields that any matrix $\wt$ satisfying the constraints of the
least squares problem must have squared Frobenius error at least
$\frobnorm{\wtmatrix - \wtstar}^2 \geq \plaincon' \numrows \numcols$,
for some universal constant $\plaincon' > 0$.


\subsubsection{Proof of Proposition~\ref{PropBreakGreedyDecomp}}

First let $\numrows = \numcols = \permrank = 2$. Consider the $(2
\times 2)$ matrix $\wt$ defined as
\begin{align*}
\wt \defn 
\begin{bmatrix}
0 & .6 \\ .6 & .4
\end{bmatrix}.
\end{align*}
It is easy to verify that $\wt \in \permset(2) \backslash
\permset(1)$.

Let us now investigate the operation of the proposed algorithm on this
matrix $\wt$. The following lemma controls the first step of the
algorithm.
\begin{lemma}
  \label{LemGreedyDecompStep1}
When the input matrix $\wt$ is as defined above, the algorithm will
select
\begin{align*}
\wthat^{(1)} =
\begin{bmatrix}
0 & .4 \\ .4 & .4
\end{bmatrix}
\end{align*}
in the first iteration.
\end{lemma}

As a consequence of this lemma, we have the following residual that is
used for the subsequent iterations of the algorithm:
\begin{align*}
\wt - \wthat^{(1)} = 
\begin{bmatrix}
0 & .2 \\ .2 & 0
\end{bmatrix}.
\end{align*}
It is easy to see the that the residual matrix $(\wt - \wthat^{(1)}) \in \permset(2)
\backslash \permset(1)$. Also observe that in each iteration, the algorithm subtracts out a matrix in
$\permset(1)$ from the residual. Consequently, the algorithm will
require at least two more iterations to terminate. 
The algorithm thus necessarily outputs a decomposition with $\widehat{\permrank} \geq 3$, as claimed.

Next we extend these arguments to any arbitrary
values of $\numrows \geq 2, \numcols \geq 2, \permrank \geq
2$. Consider matrix $\wt$ with entries:
\begin{itemize}
	\itemsep0em
\item $\wt_{11} = 0$, $\wt_{12} = \wt_{21} = .6$, $\wt_{22} = .4$
\item $\wt_{ii} = 1$ for every $i \in \{3, \ldots, \permrank\}$
\item $\wt_{ij} = 0$ for every other coordinate $(i,j)$.
\end{itemize} 
The matrix $\wt$ has a block-diagonal structure with the top-left $(2
\times 2)$ block as one non-zero component and $(\permrank - 2)$ other
entries on the diagonal as $(\permrank - 2)$ additional non-zero
components. 

The rest of the proof is partitioned into two cases:\\
\noindent \underline{Case I:} Suppose that at some step $\widehat{\permrank}$ of the algorithm, some entry of the residual matrix $(\wt - \sum_{\ell=1}^{\widehat{\permrank}} \wthat^{(\ell)}$ is strictly negative. Then  the algorithm will never terminate because every subsequent candidate matrix in the minimization step of the algorithm must lie in $\permone$ and hence must have non-negative entries. The algorithm will thus output $\widehat{\permrank} = \infty$.\\

\noindent \underline{Case II:} Now suppose that the residual matrices always have non-negative entires. Then given the block-diagonal structure of $\wt$, any matrix $\wthat^{(1)}, \wthat^{(2)},\ldots$ in the iterations of the algorithm can be
non-zero in exactly one of these diagonal components. As a result, the overall decomposition yielded
by the algorithm decouples into $\permrank$ individual decompositions
of the $\permrank$ respective blocks, each of which will contribute a
permutation-rank of at least $1$. Moreover, from the arguments for the
case of $\numrows = \numcols = 2$ above, we also have that the
top-left $(2 \times 2)$ block will induce a decomposition of
permutation-rank $3$ in the algorithm. Putting the pieces together, we
see that the matrix $\wt$ will induce the proposed algorithm to output
a decomposition of permutation-rank at least $(\permrank + 1)$,
whereas $\fnpermrank{\wt} = \permrank$.


\paragraph*{Proof of Lemma~\ref{LemGreedyDecompStep1}}

Since $\wt_{11} = 0$, we must have $\wthat^{(1)}_{11} = 0$. Now
suppose the column ordering of $\wthat^{(1)}$ is such that the first
column is greater than the second column. Then we must have
$\wthat^{(1)}_{12} = 0$. Since $\wt^{(1)}$ is the minimizer of the
optimization program in the algorithm we must then have
$\wthat^{(1)}_{21} = \wt_{21}$ and $\wthat^{(1)}_{22} = \wt_{22}$, and
consequently $\matsnorm{\wt - \wthat^{(1)}}{\entrynorm} =
\wthat^{(1)}_{12} = .6$. An analogous argument holds if the first row
is greater than the second row in the permutation of
$\wthat^{(1)}$. Finally, suppose that in the permutations of
$\wt^{(1)}$, the second column is greater than the first column and
the second row is greater than the first row. Then we must have $.4
\geq \wthat^{(1)}_{22} \geq \max\{\wthat^{(1)}_{12}, \wthat^{(1)}_{21}
\}$. With this condition, one can see that the minimizer of the
optimization program is $\wthat^{(1)}_{12} = \wthat^{(1)}_{21} =
\wthat^{(1)}_{22} = .4$. Consequently, we have $\matsnorm{\wt -
  \wthat^{(1)}}{\entrynorm} = .2 \times 2^{\frac{1}{\entrynorm}} <
.6$  for any $q \geq 1$. Thus the
algorithm chooses
\begin{align*}
\wthat^{(1)} = 
\begin{bmatrix}
0 & .4 \\ .4 & .4
\end{bmatrix}.
\end{align*}


\section{Alternative interpretation of the non-negative rank model}
\label{SecAlternativeInterpretationOfNNRank}

In the non-negative rank model described in the introduction, one may
wonder why the affinity of a user to a movie conditioned on a feature
must be modeled as the product $\lowLeftVec{\ell}_i
\lowRightVec{\ell}_j$ of the separate connections of the user and
movie to the feature. Secondly, one may also wonder why the net
affinity of a user to a movie is the sum of the affinities across the
features $\sum_{\ell = 1}^{\nnrank} \lowLeftVec{\ell}_i
\lowRightVec{\ell}_j$. These two modeling assumptions may sometimes be
confusing, and hence in what follows, we present an alternative
interpretation of the low non-negative rank model for the recommender
systems application.

Consider any feature $\ell \in [\nnrank]$. The affinities of users
towards movies conditioned on this feature is a matrix, say
$X^{(\ell)} \in [0,1]^{\numrows \times \numcols}$. The matrix
$X^{(\ell)}$ is assumed to have a (non-negative) rank of $1$. Hence
the probability that user $i$ likes movie $j$, when asked to judge
only based on feature $\ell$, equals $X^{(\ell)}_{ij}$.

Now, every user is assumed to have their own way of weighing
features to decide which movies they like. Specifically, any user
$i \in [\numrows]$ is associated to values
$\alpha_i^{(1)},\ldots,\alpha_i^{(\nnrank)}$ such that
$\alpha_i^{(\ell)} \geq 0$ for every $\ell \in [\nnrank]$ and
$\sum_{\ell=1}^{\nnrank} \alpha_i^{(\ell)} = 1$. The probability that
user $i$ likes any movie $j$ is assumed to be the convex combination
\begin{align*}
\sum_{\ell=1}^{\nnrank} \alpha_i^{(\ell)} X^{(\ell)}_{ij}.
\end{align*}

This completes the description of the model. 

Let us verify that the
resulting user-movie matrix has a non-negative rank of $\nnrank$. Recall the assumption that $X^{(\ell)}$ has a non-negative
rank of $1$, and let $X^{(\ell)} = \lowLeftVec{\ell}
(\lowRightVec{\ell})^T$ for some vectors $\lowLeftVec{\ell}$ and
$\lowRightVec{\ell}$. Then the $i^{th}$ row of the overall user-movie
matrix equals \mbox{$\sum_{\ell=1}^{\nnrank} \alpha_i^{(\ell)}
  \lowLeftVec{\ell}_i (\lowRightVec{\ell})^T$}, and hence the overall
user-movie matrix equals
\begin{align*}
\sum_{\ell=1}^{\nnrank} \lowLeftVecTilde{\ell} (\lowRightVec{\ell})^T,
\qquad \mbox{where } \quad \lowLeftVecTilde{\ell} =
\begin{bmatrix}
\alpha_1^{(\ell)} \lowLeftVec{\ell}_1 \\
\vdots \\
\alpha_\numrows^{(\ell)} \lowLeftVec{\ell}_\numrows
\end{bmatrix}.
\end{align*}
This completes the alternative description of the non-negative rank
model.

One can observe that the restriction $\sum_{\ell=1}^{\nnrank}
\alpha_i^{(\ell)} = 1$ makes this model slightly more restrictive than
the non-negative rank model described earlier in the main
text. However, all of our results on
estimation for the non-negative rank model described in Section~\ref{SecStat} continue to apply to this
model as well.


\let\oldbibliography\thebibliography
\renewcommand{\thebibliography}[1]{\oldbibliography{#1}
	\setlength{\itemsep}{0pt}}

\bibliographystyle{alpha_initials} 
\bibliography{bibtex}

\newcommand{\etalchar}[1]{$^{#1}$}
\begin{thebibliography}{VHVVD08}

\bibitem[AGKM12]{arora2012computing}
S. Arora, R. Ge, R. Kannan, and A. Moitra.
\newblock Computing a nonnegative matrix factorization--provably.
\newblock In {\em Proceedings of the forty-fourth annual ACM symposium on
  Theory of computing}, pages 145--162. ACM, 2012.

\bibitem[BZ16]{balcan2016noise}
M.-F.~F. Balcan and H. Zhang.
\newblock Noise-tolerant life-long matrix completion via adaptive sampling.
\newblock In {\em Advances In Neural Information Processing Systems}, 2016.

\bibitem[CCS10]{cai2010singular}
J.-F. Cai, E.~J. Cand{\`e}s, and Z. Shen.
\newblock A singular value thresholding algorithm for matrix completion.
\newblock {\em SIAM Journal on Optimization}, 20(4):1956--1982, 2010.

\bibitem[CGMS17]{chen2017competitive}
X. Chen, S. Gopi, J. Mao, and J. Schneider.
\newblock Competitive analysis of the top-k ranking problem.
\newblock In {\em ACM-SIAM Symposium on Discrete Algorithms}, 2017.

\bibitem[Cha14]{chatterjee2014matrix}
S. Chatterjee.
\newblock Matrix estimation by universal singular value thresholding.
\newblock {\em The Annals of Statistics}, 43(1):177--214, 2014.

\bibitem[CJSC13]{chen13}
Y. Chen, A. Jalali, S. Sanghavi, and C. Caramanis.
\newblock Low-rank matrix recovery from errors and erasures.
\newblock {\em {IEEE} Trans. Information Theory}, 59(7):4324--4337, 2013.

\bibitem[CM16]{chatterjee2016estimation}
S. Chatterjee and S. Mukherjee.
\newblock On estimation in tournaments and graphs under monotonicity
  constraints.
\newblock {\em arxiv:1603.04556}, 2016.

\bibitem[CO10]{cai2010fast}
J.-F. Cai and S. Osher.
\newblock Fast singular value thresholding without singular value
  decomposition.
\newblock {\em UCLA CAM Report}, 5, 2010.

\bibitem[CR09]{candes09exact}
E.~J. Candes and B. Recht.
\newblock Exact matrix completion via convex optimization.
\newblock {\em Found. Comput. Math.}, 9(6):717--772, December 2009.

\bibitem[CT10]{candes10power}
E.~J. Cand\`{e}s and T. Tao.
\newblock The power of convex relaxation: Near-optimal matrix completion.
\newblock {\em IEEE Trans. Inf. Theor.}, 56(5):2053--2080, May 2010.

\bibitem[CT12]{cover2012elements}
T. Cover and J. Thomas.
\newblock {\em Elements of information theory}.
\newblock John Wiley \& Sons, 2012.

\bibitem[DG{\etalchar{+}}14]{donoho2014minimax}
D. Donoho, M. Gavish, et~al.
\newblock Minimax risk of matrix denoising by singular value thresholding.
\newblock {\em The Annals of Statistics}, 42(6):2413--2440, 2014.

\bibitem[DR16]{davenport2016overview}
M.~A. Davenport and J. Romberg.
\newblock An overview of low-rank matrix recovery from incomplete observations.
\newblock {\em IEEE Journal of Selected Topics in Signal Processing},
  10(4):608--622, 2016.

\bibitem[DS03]{donoho2003does}
D. Donoho and V. Stodden.
\newblock When does non-negative matrix factorization give a correct
  decomposition into parts?
\newblock In {\em Advances in neural information processing systems}, 2003.

\bibitem[FMR16]{flammarion2016optimal}
N. Flammarion, C. Mao, and P. Rigollet.
\newblock Optimal rates of statistical seriation.
\newblock {\em arxiv:1607.02435}, 2016.

\bibitem[Gil52]{gilbert1952comparison}
E.~N. Gilbert.
\newblock A comparison of signalling alphabets.
\newblock {\em Bell System Technical Journal}, 31(3):504--522, 1952.

\bibitem[Gil12]{gillis2012sparse}
N. Gillis.
\newblock Sparse and unique nonnegative matrix factorization through data
  preprocessing.
\newblock {\em Journal of Machine Learning Research}, 13(Nov):3349--3386, 2012.

\bibitem[Gil14]{gillis2014and}
N. Gillis.
\newblock The why and how of nonnegative matrix factorization.
\newblock {\em Regularization, Optimization, Kernels, and Support Vector
  Machines}, 12(257), 2014.

\bibitem[GPH04]{gobinet2004application}
C. Gobinet, E. Perrin, and R. Huez.
\newblock Application of non-negative matrix factorization to fluorescence
  spectroscopy.
\newblock In {\em European Signal Processing Conference}, 2004.

\bibitem[Gro11]{gross11}
D. Gross.
\newblock Recovering low-rank matrices from few coefficients in any basis.
\newblock {\em {IEEE} Trans. Information Theory}, 57(3):1548--1566, 2011.

\bibitem[H{\aa}s90]{haastad1990tensor}
J. H{\aa}stad.
\newblock Tensor rank is np-complete.
\newblock {\em Journal of Algorithms}, 11(4):644--654, 1990.

\bibitem[Hit27]{hitchcock1927expression}
F.~L. Hitchcock.
\newblock The expression of a tensor or a polyadic as a sum of products.
\newblock {\em Studies in Applied Mathematics}, 6(1-4):164--189, 1927.

\bibitem[HK13]{hsu2013learning}
D. Hsu and S.~M. Kakade.
\newblock Learning mixtures of spherical gaussians: moment methods and spectral
  decompositions.
\newblock In {\em ACM Innovations in Theoretical Computer Science}, 2013.

\bibitem[HND15]{hsieh2015pu}
C.-J. Hsieh, N. Natarajan, and I. Dhillon.
\newblock Pu learning for matrix completion.
\newblock In {\em International Conference on Machine Learning}, pages
  2445--2453, 2015.

\bibitem[HSRW16]{heckel2016active}
R. Heckel, N.~B. Shah, K. Ramchandran, and M.~J. Wainwright.
\newblock Active ranking from pairwise comparisons and when parametric
  assumptions don't help.
\newblock {\em arxiv:1606.08842}, 2016.

\bibitem[JKN16]{jin2016provable}
C. Jin, S.~M. Kakade, and P. Netrapalli.
\newblock Provable efficient online matrix completion via non-convex stochastic
  gradient descent.
\newblock In {\em Advances in Neural Information Processing Systems}, 2016.

\bibitem[JNS13]{jain2013low}
P. Jain, P. Netrapalli, and S. Sanghavi.
\newblock Low-rank matrix completion using alternating minimization.
\newblock In {\em ACM symposium on Theory of computing}, 2013.

\bibitem[KB09]{kolda2009tensor}
T.~G. Kolda and B.~W. Bader.
\newblock Tensor decompositions and applications.
\newblock {\em SIAM review}, 51(3):455--500, 2009.

\bibitem[KBV09]{koren2009matrix}
Y. Koren, R. Bell, and C. Volinsky.
\newblock Matrix factorization techniques for recommender systems.
\newblock {\em Computer}, 42(8), 2009.

\bibitem[Klo14]{klopp2014noisy}
O. Klopp.
\newblock Noisy low-rank matrix completion with general sampling distribution.
\newblock {\em Bernoulli}, 20(1):282--303, 2014.

\bibitem[KLT11]{koltchinskii2011nuclear}
V. Koltchinskii, K. Lounici, and A.~B. Tsybakov.
\newblock Nuclear-norm penalization and optimal rates for noisy low-rank matrix
  completion.
\newblock {\em The Annals of Statistics}, pages 2302--2329, 2011.

\bibitem[KMO10]{keshavan2010matrix}
R.~H. Keshavan, A. Montanari, and S. Oh.
\newblock Matrix completion from noisy entries.
\newblock {\em Journal of Machine Learning Research}, 11(Jul):2057--2078, 2010.

\bibitem[KR{\etalchar{+}}05]{klein2005concentration}
T. Klein, E. Rio, et~al.
\newblock Concentration around the mean for maxima of empirical processes.
\newblock {\em The Annals of Probability}, 33(3):1060--1077, 2005.

\bibitem[Kru77]{kruskal1977three}
J.~B. Kruskal.
\newblock Three-way arrays: rank and uniqueness of trilinear decompositions,
  with application to arithmetic complexity and statistics.
\newblock {\em Linear algebra and its applications}, 18(2):95--138, 1977.

\bibitem[Lau01]{Lau01}
M. Laurent.
\newblock Matrix completion problems.
\newblock In {\em The Encyclopedia of Optimization}, pages 221---229. Kluwer
  Academic, 2001.

\bibitem[LCP{\etalchar{+}}08]{laurberg2008theorems}
H. Laurberg, M.~G. Christensen, M.~D. Plumbley, L.~K. Hansen, and S.~H. Jensen.
\newblock Theorems on positive data: On the uniqueness of {NMF}.
\newblock {\em Computational intelligence and neuroscience}, 2008, 2008.

\bibitem[LS99]{lee1999learning}
D.~D. Lee and H.~S. Seung.
\newblock Learning the parts of objects by non-negative matrix factorization.
\newblock {\em Nature}, 401(6755):788--791, 1999.

\bibitem[NW12]{negahban2012restricted}
S. Negahban and M.~J. Wainwright.
\newblock Restricted strong convexity and weighted matrix completion: Optimal
  bounds with noise.
\newblock {\em Journal of Machine Learning Research}, 13(May):1665--1697, 2012.

\bibitem[PWC16]{pananjady2016linear}
A. Pananjady, M.~J. Wainwright, and T.~A. Courtade.
\newblock Linear regression with an unknown permutation: Statistical and
  computational limits.
\newblock {\em arxiv:1608.02902}, 2016.

\bibitem[Rec11]{recht2011simpler}
B. Recht.
\newblock A simpler approach to matrix completion.
\newblock {\em Journal of Machine Learning Research}, 12(Dec):3413--3430, 2011.

\bibitem[SAJ05]{srebro2005generalization}
N. Srebro, N. Alon, and T.~S. Jaakkola.
\newblock Generalization error bounds for collaborative prediction with
  low-rank matrices.
\newblock In {\em Advances In Neural Information Processing Systems}, 2005.

\bibitem[SBGW17]{shah2015stochastically}
N.~B. Shah, S. Balakrishnan, A. Guntuboyina, and M.~J. Wainwright.
\newblock Stochastically transitive models for pairwise comparisons:
  Statistical and computational issues.
\newblock {\em IEEE Transactions on Information Theory}, 2017.

\bibitem[SBW16a]{shah2016feeling}
N.~B. Shah, S. Balakrishnan, and M.~J. Wainwright.
\newblock Feeling the {B}ern: Adaptive estimators for {B}ernoulli probabilities
  of pairwise comparisons.
\newblock {\em arxiv:1603.06881}, 2016.

\bibitem[SBW16b]{shah2016permutation}
N.~B. Shah, S. Balakrishnan, and M.~J. Wainwright.
\newblock A permutation-based model for crowd labeling: Optimal estimation and
  robustness.
\newblock {\em arxiv:1606.09632}, 2016.

\bibitem[Sch68]{schonemann1968two}
P.~H. Sch{\"o}nemann.
\newblock On two-sided orthogonal procrustes problems.
\newblock {\em Psychometrika}, 33(1):19--33, 1968.

\bibitem[Sha17]{shah2017thesis}
N. Shah.
\newblock {\em Learning From People}.
\newblock PhD thesis, EECS Department, University of California, Berkeley, Jul
  2017.

\bibitem[SK11]{sarver2011application}
R. Sarver and A. Klapuri.
\newblock Application of nonnegative matrix factorization to signal-adaptive
  audio effects.
\newblock In {\em Proc. DAFx}, pages 249--252, 2011.

\bibitem[SW15]{shah2015simple}
N.~B. Shah and M.~J. Wainwright.
\newblock Simple, robust and optimal ranking from pairwise comparisons.
\newblock {\em arxiv:1512.08949}, 2015.

\bibitem[Tao12]{tao2012topics}
T. Tao.
\newblock {\em Topics in random matrix theory}, volume 132.
\newblock American Mathematical Society Providence, RI, 2012.

\bibitem[TST05]{theis2005first}
F.~J. Theis, K. Stadlthanner, and T. Tanaka.
\newblock First results on uniqueness of sparse non-negative matrix
  factorization.
\newblock In {\em European Signal Processing Conference}, 2005.

\bibitem[Var57]{varshamov1957estimate}
R. Varshamov.
\newblock Estimate of the number of signals in error correcting codes.
\newblock In {\em Dokl. Akad. Nauk SSSR}, 1957.

\bibitem[VHVVD08]{vandendorpe2008parameterization}
A. Vandendorpe, N.-D. Ho, S. Vanduffel, and P. Van~Dooren.
\newblock On the parameterization of the creditrisk+ model for estimating
  credit portfolio risk.
\newblock {\em Insurance: Mathematics and Economics}, 42(2):736--745, 2008.

\bibitem[YLP15]{yun2015streaming}
S.-Y. Yun, M. Lelarge, and A. Proutiere.
\newblock Streaming, memory limited matrix completion with noise.
\newblock {\em arxiv:1504.03156}, 2015.

\end{thebibliography}

\end{document}